\documentclass[10pt,twocolumn,letterpaper]{article}

\usepackage{iccv}
\usepackage{times}
\usepackage{epsfig}
\usepackage{graphicx}
\usepackage{amsmath}
\usepackage{amssymb}

\usepackage{booktabs}

\usepackage[utf8]{inputenc} % allow utf-8 input
\usepackage[T1]{fontenc}    % use 8-bit T1 fonts
\usepackage{url}            % simple URL typesetting
\usepackage{colortbl}
\usepackage{amsfonts}       % blackboard math symbols
\usepackage{nicefrac}       % compact symbols for 1/2, etc.
\usepackage{microtype}      % microtypography
\usepackage[dvipsnames]{xcolor}         % colors
\usepackage{pifont} 
\usepackage{mwe}
\usepackage{enumitem}
\usepackage{dsfont}

\usepackage{multirow}
\usepackage[ruled, lined, linesnumbered, commentsnumbered, longend]{algorithm2e}
\usepackage{xcolor}
\usepackage{gensymb}
\usepackage{wrapfig}
\usepackage{floatrow}

\usepackage[accsupp]{axessibility}  % Improves PDF readability for those with disabilities.

\definecolor{BlueArrow}{rgb}{0.0,0.463,0.729}
\definecolor{RedArrow}{rgb}{0.710,0.090,0.0}
\definecolor{RedArrowChris}{rgb}{0.77,0.35,0.07}
\definecolor{MapNavigable}{rgb}{1,0.83,0.47}
\definecolor{MapObstacles}{rgb}{0.45,0.98,0.48}
\definecolor{MapUnexplored}{rgb}{0.55,0.84,1}
\definecolor{MidRuleGray}{rgb}{0.7,0.7,0.7}

\newcommand{\myparagraph}[1]{
\textbf{#1} ---
}
\usepackage[pagebackref=true,breaklinks=true,letterpaper=true,colorlinks,bookmarks=false]{hyperref}

\iccvfinalcopy % *** Uncomment this line for the final submission

\def\iccvPaperID{1059} % *** Enter the ICCV Paper ID here

\ificcvfinal\fi

\begin{document}

%%%%%%%%% TITLE
\title{Multi-Object Navigation with dynamically learned neural implicit representations}

\author{
  Pierre Marza $^1$ \qquad
  Laetitia Matignon$^2$ \qquad
  Olivier Simonin$^1$ \qquad
  Christian Wolf$^3$ \qquad
  \\
  {\normalsize$^1$INSA Lyon} \quad
  {\normalsize$^2$UCBL} \quad
  {\normalsize$^3$Naver Labs Europe}  \quad
  \\
  {\tt \small \{pierre.marza, olivier.simonin\}@insa-lyon.fr} \\
  {\tt \small laetitia.matignon@univ-lyon1.fr}, {\tt \small christian.wolf@naverlabs.com} \\
  {\normalsize Project Page: \href{https://pierremarza.github.io/projects/dynamic_implicit_representations/}{https://pierremarza.github.io/projects/dynamic\_implicit\_representations/}}
}

\maketitle
% Remove page # from the first page of camera-ready.
\ificcvfinal\thispagestyle{empty}\fi

%%%%%%%%% ABSTRACT
\begin{abstract}
\noindent
   Understanding and mapping a new environment are core abilities of any autonomously navigating agent. While classical robotics usually estimates maps in a stand-alone manner with SLAM variants, which maintain a topological or metric representation, end-to-end learning of navigation keeps some form of memory in a neural network. Networks are typically imbued with inductive biases, which can range from vectorial representations to birds-eye metric tensors or topological structures. In this work, we propose to structure neural networks with two neural implicit representations, which are learned dynamically during each episode and map the content of the scene: (i) the \textit{Semantic Finder} predicts the position of a previously seen queried object; (ii) the \textit{Occupancy and Exploration Implicit Representation} encapsulates information about explored area and obstacles, and is queried with a novel global read mechanism which directly maps from function space to a usable embedding space. Both representations are leveraged by an agent trained with Reinforcement Learning (RL) and learned online during each episode. We evaluate the agent on Multi-Object Navigation and show the high impact of using neural implicit representations as a memory source.
\end{abstract}

%%%%%%%%% BODY TEXT
\begin{figure}[t] \centering
    \includegraphics[width=0.9\linewidth]{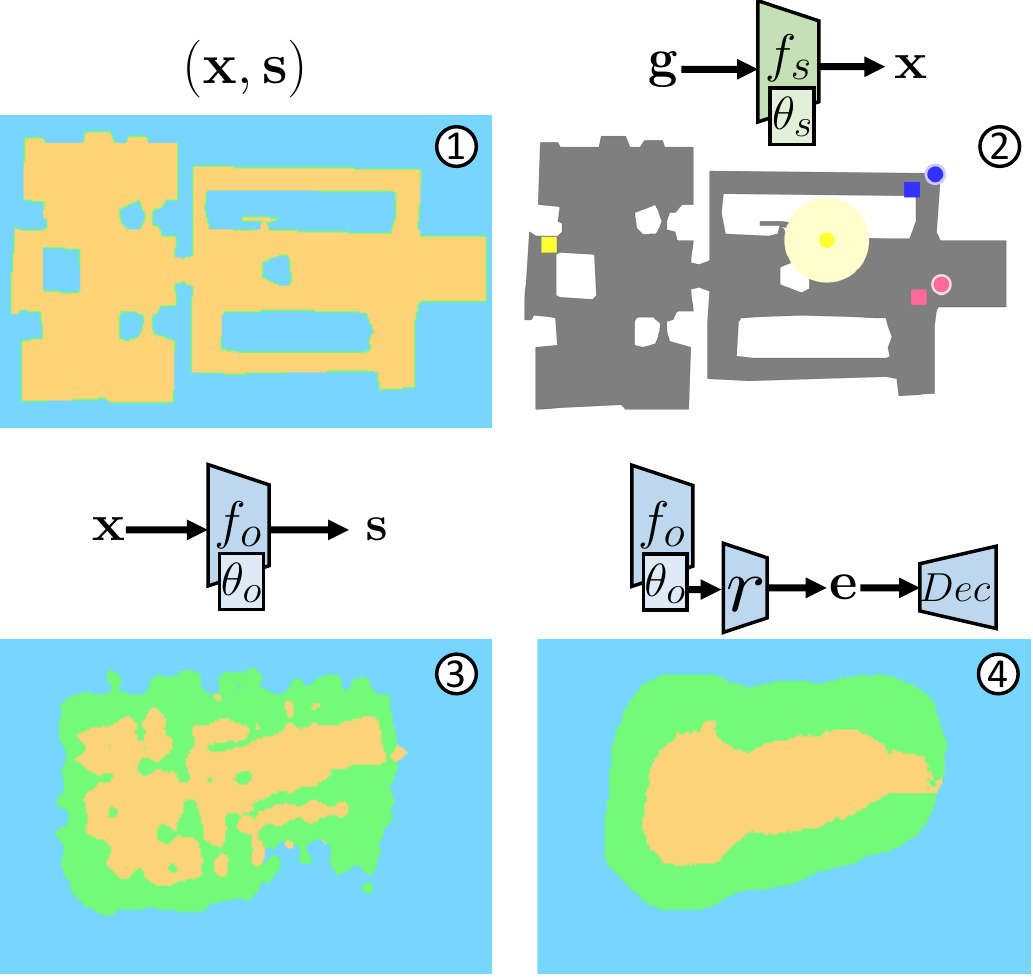} 
    \caption{\label{fig:teaser}We propose \textbf{two implicit representations} as inductive biases for autonomous agents --- both are learned online during each episode. {\large \ding{193}} a semantic representation $f_s$ predicts positions $\mathbf{x}$ from goals $\mathbf{g}$ given as semantic codes. We show Ground Truth object positions (rectangles) and predictions (round; radius shows uncertainty, unit-less, as an illustration). Blue and pink objects have been observed, but not the yellow target. {\large\ding{194}} A structural representation $f_o$ predicts occupancy and exploration $\mathbf{s}$ from positions  $\mathbf{x}$; we provide a global read which directly maps from function space $f_o$ (represented by trainable weights $\theta_o$) to a context embedding $\mathbf{e}$ used by the agent. {\large\ding{195}} shows the reconstruction produced by a decoder $Dec$ during training. \textcolor{MapNavigable}{Orange=navigable}, \textcolor{MapObstacles}{Green=Obstacles}, \textcolor{MapUnexplored}{Blue=Unexplored}. {\large\ding{192}} a ground-truth map is shown for reference, simulating a fully explored scene.} 
\end{figure}

\section{Introduction}
\noindent
Autonomous navigation in complex unknown 3D environments from visual observations requires building a suitable representation of the environment, in particular when the targeted navigation task requires high-level reasoning. Whereas classical robotics builds these representations explicitly through reconstructions, possibly supported through machine learning, end-to-end training learns them automatically either from reward, by imitation learning or through self-supervised objectives.

While spatial representations can emerge even in unstructured agents, as shown in the form of grid-cells in artificial \cite{GridCellsICLR2018,GridCellsDeepmind2018} and biological agents \cite{haftinggridcell}, spatial inductive biases can support learning actionable spatial representations and decrease sample complexity. Popular inductive biases are metric maps ~\cite{DBLP:conf/iclr/ParisottoS18, DBLP:conf/pkdd/BeechingD0020, Henriques_2018_CVPR},  topological maps~\cite{beeching2020learning, Chaplot_2020_CVPR} and recently, self-attention, adapting transformers~\cite{vaswani2017attention} to sequential decision making and navigation \cite{Fang_2019_CVPR,du_vtnet_2021,chen_think_2022,reed_generalist_2022}. 
The chosen representation should support robust estimation of navigable space even in difficult conditions, mapping features and objects of interest, as well as querying and reusing this information at a later time. The representation should be as detailed as required, span the full (observed) scene, easy to query, and efficient to read  and write to, in particular when training is done in large-scale simulations.

Our work builds on neural fields and implicit representations, a category of models which represent the scene geometry, and eventually the semantics, by the weights of a trained neural network~\cite{xie2021neural}. They have the advantage of avoiding the explicit choice of scene representation (e.g. volume, surface, point cloud etc.) and inherently benefit from the generalization abilities of deep networks to interpolate and complete unobserved information. Implicit representations have demonstrated impressive capabilities in novel view synthesis~\cite{mildenhall2020nerf, sitzmann2020implicit}, and have potential as a competitive representation for robotics \cite{ortiz_isdf_2022,li_learning_2022,sucar2021imap,adamkiewicz2022vision}. Their continuous nature allows them to handle level of detail efficiently through a budget given as the amount of trainable weights. This allows to span large environments without the need of discretizing the environment and handling growing maps. 

We explore and study the potential of implicit representations as inductive biases for visual navigation. Similar to recent work in implicit SLAM~\cite{sucar2021imap}, our representations are dynamically learned in each episode. Going beyond, we exploit the representation dynamically in one of the most challenging visual navigation tasks, \textit{Multi-Object Navigation}~\cite{DBLP:conf/nips/WaniPJCS20}. We introduce two complementary representations, namely a query-able \textit{Semantic Finder} trained to predict the scene coordinates of an object of interest specified as input, and an \textit{Occupancy and Exploration Implicit Representation}, which maps 2D coordinates to occupancy information, see Figure \ref{fig:teaser}. We address the issue of the efficiency of querying an implicit representation globally by introducing a new global read mechanism, which directly maps from function space, represented through its trainable parameters, to an embedding  summarizing the current status of occupancy and exploration information, useful for navigation. Invariance w.r.t. reparametrization of the queried network is favored (but not enforced) through a transformer based solution. Our method does \textit{not} require previous rollouts on the scene for pre-training or building a representation.

Our work targets a fundamental aspect of visual and semantic navigation, the mapping of key objects of interest. \textit{MultiON} is currently one of the few benchmarks which evaluates it. As argued in previous literature \cite{DBLP:conf/pkdd/BeechingD0020,DBLP:conf/nips/WaniPJCS20}, only sequential tasks, where objects have to be found in a given order, allow object level mapping to emerge directly from reward. This follows from the observation that an agent trained to find and retrieve a single object per episode (from reward) is not required to map seen target objects, as observing them directly leads to a reactive motion towards them.

Our contributions can be summarized as follows:
(i) We propose two implicit representations for semantic, occupancy and exploration information, which are trained online during each episode;
(ii) We introduce a new global read procedure which can extract summarizing context information directly from the function itself;
(iii) We show that the representations obtain performance gains compared to  classical neural agents;
(iv) We evaluate and analyze key design choices, the representation's scaling laws and its capabilities of lifelong learning.

\section{Related Work}
\myparagraph{Visual Navigation} is a rich problem that involves perception, mapping and decision making, with required capacities being highly dependent on the specific task. A summary of reasoning in navigation has been given in~\cite{DBLP:journals/corr/abs-1807-06757}, differentiating, for instance, between waypoint navigation 
(\textit{Pointgoal})~\cite{DBLP:journals/corr/abs-1807-06757} or finding objects of semantic categories (\textit{ObjectGoal}) ~\cite{DBLP:journals/corr/abs-1807-06757}. More recent tasks have been explicitely designed to evaluate and encourage mapping objects of interest during navigation itself \cite{DBLP:conf/icpr/BeechingD0020, DBLP:conf/nips/WaniPJCS20}. 
They are of sequential nature and  use external objects, which are not part of the scanned 3D scenes but randomly placed. In this work we address \textit{Multi-Object Navigation (MultiON)}~\cite{DBLP:conf/nips/WaniPJCS20}.

\myparagraph{Mapping and Representations} Classical methods often rely on SLAM~\cite{bresson2017simultaneous,lluvia_active_2021} which has been proposed in different variants (2D or 3D metric, topological, eventually with semantics) and observations (LIDAR, visual etc.). The objective is to integrate observations and odometry estimates over a trajectory and reconstruct the scene. Differentiable variants have been proposed recently \cite{jatavallabhula_gradslam_2020,karkus_differentiable_2021}.
Mapping can also be discovered through interactions by a blind agent \cite{MoleArxiv2022}.
Visual Navigation can also be framed as an end-to-end learning problem, where representations are learned automatically from different signals, in particular RL. Memory can take the form of vectorial representations in recurrent units~\cite{DBLP:conf/iclr/MirowskiPVSBBDG17, DBLP:conf/iclr/JaderbergMCSLSK17, DBLP:conf/icra/ZhuMKLGFF17}, with hybrid variants including mapping \cite{Chaplot_2020_CVPR,AssemICRA2023,DeyIROS2023}.
Recent work tends to augment agents with more structured memories. Examples are spatial metric tensors, which can contain occupancy~\cite{Chaplot2020Learning}, semantics~\cite{chaplot2020object} or be fully latent, effectively corresponding to inductive biases of the neural agents~\cite{DBLP:conf/iclr/ParisottoS18, DBLP:conf/pkdd/BeechingD0020, Henriques_2018_CVPR}. Other alternatives are topological maps~\cite{beeching2020learning, Chaplot_2020_CVPR} or self-attention and transformers~\cite{vaswani2017attention} adapted to navigation~\cite{Fang_2019_CVPR,du_vtnet_2021,chen_think_2022,reed_generalist_2022}.

\myparagraph{Implicit representations} were initially targeting 3D reconstruction~\cite{mescheder2019occupancy, park2019deepsdf, chen2019learning}. The core idea is to replace the need for discretizing 3D space into voxels~\cite{maturana2015voxnet}, 3D points~\cite{fan2017point} or meshes~\cite{groueix2018papier}, by an implicit representation of the 3D structure of the scene through the parameters of a learned neural network. Recent work~\cite{mildenhall2020nerf, sitzmann2020implicit} achieved state-of-the-art performance on novel view synthesis with neural implicit representations. The NeRF paper introduced a differentiable volume rendering loss allowing to supervise 3D scene reconstruction from only 2D supervision~\cite{mildenhall2020nerf}. For a more detailed overview of recent advances in the rapidly growing field, we refer the reader to~\cite{xie2021neural}.

\myparagraph{Implicit representations in robotics} are a recent phenomenon, used to represent density~\cite{adamkiewicz2022vision} or to perform visuomotor control~\cite{li20223d}. Related to goal-oriented navigation, some work targets SLAM with neural implicit representations~\cite{sucar2021imap}, follow-up adding semantics~\cite{zhi2021place}, learned from sparse semantic annotations of the scene. \cite{zhi2021ilabel} is also built on top of~\cite{sucar2021imap} and allows a user to interactively provide semantic annotation for the implicit representation to be trained on in real time. \cite{zhu2021nice} proposes a hierarchical implicit representation of a scene to scale to larger environments and obtain a more detailed reconstruction. \cite{chung2023orbeez} combines feature-based SLAM and NeRF. Our work goes beyond implicit SLAM and does not stop at reconstructing a scene. We not only build implicit representations dynamically during the episode, we also use them in a down-stream navigation task \textit{without} requiring any initial rollout for pre-training or building a representation. We also combine two different implicit representations targeting semantics vs. scene structure.

\myparagraph{Analyzing the neural network function space} implicit representations are instances of function spaces, which are represented through their trainable parameters. Previous work performed analyses by predicting accuracy from network weights~\cite{unterthiner2020predicting, martin2020heavy, martin2021predicting} or the generality gap between train and test performance from hidden activations~\cite{jiang2018predicting, yak2019towards}. A direction pioneered by Hypernetworks~\cite{ha2016hypernetworks}  directly predict the network weights. Recently, \cite{zhmoginov2022hypertransformer} generate the weights of a CNN from support samples in the context of few-shot learning. More related to our work, \cite{pan2022neural} learns to predict the weights of an implicit representation based on external factors in the context of spatio-temporal dynamics encoding.
In this work, we learn a direct mapping between an implicit representation, represented by its weights, to an actionable embedding summarizing the scene globally.

\section{Navigating with implicit representations}
\noindent
We target the \textit{Multi-Object Navigation} task ~\cite{DBLP:conf/nips/WaniPJCS20}, which requires an agent to navigate in a photo realistic 3D environment from RGB-D observations $\mathbf{o}_t$ and reach a sequence of target objects (colored cylinders) in a particular order. Goal categories $\mathbf{g}_t$ are given at each time step $t$. In an RL setting, the agent receives positive reward for each successfully reached object as well as when the geodesic distance towards the goal decreases, and a small negative reward for each step, favoring short paths.

\begin{figure*}
  \label{fig:high-level}
  \centering
  \includegraphics[width=0.8\textwidth]{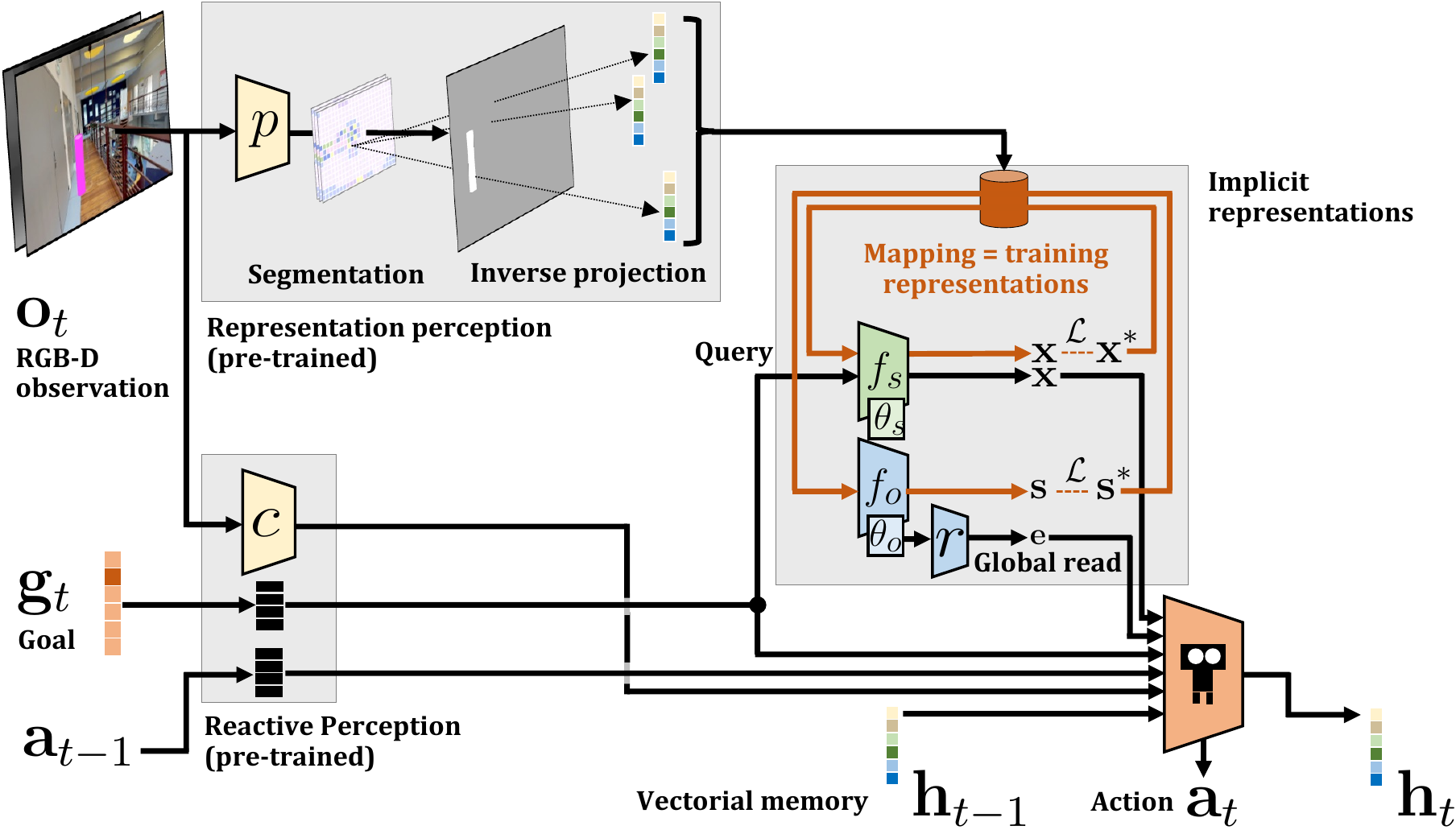}
  \caption{\label{fig:overview}\textbf{Navigating with implicit representations}. Red connections \includegraphics[height=0.5em]{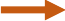} indicate the  training process of the two implicit representations (=mapping), which is also done during agent deployment. Black connections \includegraphics[height=0.5em]{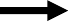} show the forward pass of the agent. \includegraphics[height=0.8em]{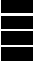} are discrete learned embeddings (LUT). Policy training is not shown in this figure.}
\end{figure*}

We follow and augment a base end-to-end architecture used in many recent RL approaches, including ~\cite{DBLP:conf/nips/WaniPJCS20, marza2021teaching, DBLP:conf/iclr/MirowskiPVSBBDG17}, with the RGB-D observation, class of the current target and previous action as input to the agent. Temporal information is aggregated with a GRU unit whose output is fed to actor and critic heads. 
We equip this agent with two implicit representations, trained to hold and map essential information necessary for navigation: the positions of different objects of interest, and occupancy / exploration information, as shown in Figure \ref{fig:teaser},

\begin{itemize}[leftmargin=8mm]
\item[{\includegraphics[height=1em]{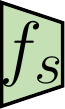}}]
The goal of the \textit{Semantic Finder} $f_s(.;\theta_s)$ parameterized by trainable weights $\theta_s$ is to predict the absolute position of an object as $  \mathbf{x} = [ \mathbf{x}_x \ \mathbf{x}_y \ \mathbf{x}_z ] =  f_s(\mathbf{q};\theta_s)$ specified through an input query vector $\mathbf{q}$. Uncertainty $u$ is also estimated --- see Section \ref{sec:semanticfinder} for details. $\mathbf{x}$ is then converted into coordinates relative to the agent to be fed to the GRU. Compared to classical metric representations \cite{Henriques_2018_CVPR,DBLP:conf/iclr/ParisottoS18,Chaplot_2020_CVPR,DBLP:conf/pkdd/BeechingD0020}, querying the location of an object can be done through a single forward pass.

\item[{\includegraphics[height=1em]{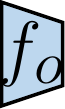}}]
The \textit{Occupancy and Exploration Representation} $f_o(.;\theta_o)$ parameterized by trainable weights $\theta_o$ encodes information about free navigable space and obstacles. It predicts 
occupancy $\mathbf{s}$ as a classification problem with three classes \textit{\{Obstacle, Navigable, Unexplored\}}, as $\mathbf{s} = f_o(\phi;\theta_o)$, where $\phi$ is a position feature vector encoded from coordinates $\mathbf{x}$ --- see Section \ref{sec:occupancyrepresentation} for details. 

\end{itemize}

The \textit{Occupancy and Exploration Representation} can in principle be queried directly for a single position, but reading out information over a large area directly this way would require multiple reads. We propose to compress this procedure by providing a trainable global read operation $r(.;\theta_g)$, which predicts an embedding $\mathbf{e}$ containing a global context about what has already been explored, and positions of navigable space. The prediction is done directly from the trainable parameters of the implicit representation, as $\mathbf{e} = r(\theta_o;\theta_r)$. Here $\theta_o$ is input to $r$, whereas $\theta_r$ are its parameters.

Given representations $f_s$ and $f_o$, a single forward pass of the agent at time step $t$ and for a goal $g_t$ involves reading the representations and providing the input to the policy. The current RGB-D observation $o_t$ is also encoded by the convolutional network $c$ (different from the projection module $p$ used to generate samples for training the \textit{Semantic Finder}). Previous action $a_{t-1}$ and current goal $g_t$ are passed through embedding layers, named $L(.)$ in the following equations. These different outputs are fed to the policy,
\vspace*{-1mm}
\begin{align}
    \label{eq:perception}
    \mathbf{x}_t &=  f_s(g_t;\theta_{s,t}), \ 
    \mathbf{e}_t = r(\theta_{o,t};\theta_r), \
    \mathbf{c}_t = c(o_t; \theta_{c}), \\
    \mathbf{h}_t &= GRU(\mathbf{h}_{t-1}, \mathbf{x}_t, u_t, \mathbf{e}_t, L(a_{t-1}), L(g_t), \mathbf{c}_t; \theta_{G}),\\
    a_t &= \pi(\mathbf{h}_t;\theta_{\pi}), 
\end{align}
where we added indices $\cdot_t$ to relevant variables to indicate time. Please note that the trainable parameters $\theta_{s,t}$ and $\theta_{o,t}$ of the two implicit representations are time dependent, as they depend on the observed scene and are updated dynamically, whereas the parameters of the policy $\pi$ and the global reader $r$ are not. Here, GRU corresponds to the update equations of a GRU network, where we omitted gates for ease of notation. The notation $a_t=\pi(.)$ predicting action $a_t$ is also a simplification, as we train the agent with PPO, an actor-critic method --- see Section \ref{sec:trainingagent}.

\myparagraph{Mapping means training!} The implicit representations $f_s$ and $f_o$ maintain a compact and actionable representation of the observed scene, and as such need to be updated at each time step from the current observation $o_t$. Given their implicit nature and implementation as neural networks, updates are gradient based and done with SGD. The implicit representations are therefore \textit{trained from scratch at each episode even after deployment}.

Training a representation from observations obtained sequentially during an episode also raises a serious issue of catastrophic forgetting \cite{Catastrophic2014}, as places of the scene observed early might be forgotten later in the episode \cite{sucar2021imap,zhi2021place}. We solve this by maintaining two replay buffers throughout the episode, one for each representation. Training samples are generated from each new observation and added to the replay buffers at each time step.
Both representations are then trained for a number of gradient steps ($n_s$ for the \textit{Semantic Finder} and $n_o$ for the \textit{Exploration and Occupancy Representation}). Details on the two representations and their training are given in Sections \ref{sec:semanticfinder} and \ref{sec:occupancyrepresentation}. The global reader $r$ is not trained or finetuned online but rather trained once offline.

\subsection{The Semantic Finder $f_s$}
\label{sec:semanticfinder}
\noindent
While recent work on implicit representations for robotics focused on signed distance functions \cite{ortiz_isdf_2022,li_learning_2022}, occupancy \cite{sucar2021imap} or density, assuming light density approximates mass density \cite{adamkiewicz2022vision}, the aim of this model is to localize an object of interest within the scene, which can be seen as inverse operation to classical work.
From a query vector given as input,  the \textit{Semantic Finder} predicts the position of the object, which is particularly useful for an agent interacting with an environment in the context of a goal conditioned task. It is implemented as a 3-layer MLP with ReLu activations in the intermediate layers and a sigmoid activation for the output.  Hidden layers have $512$ neurons. The query vector $\mathbf{q}$ corresponds to the 1-in-K encoding of the target object class, which during navigation is directly determined by the object goal $g_t$ provided by the task.

\myparagraph{Mapping/Training} The implicit representation is updated
minimizing the L1 loss between the prediction $\mathbf{x}_i = f_s(\mathbf{q}_i,\theta_s)$ and the supervised coordinates $\mathbf{x}_i^*$ (we avoided the term ``\textit{ground-truth}'' here on purpose), 
$
\mathcal{L}_s = \sum_i \|\mathbf{x}_i^* - \mathbf{x}_i\|_{1}
$, where the sum goes over the batch sampled from the scene replay buffer. Coordinates $\mathbf{x}_i^*$ are normalized  $\in[0,1]$.

The data pairs $(x_i^*, q_i)$ for training are created from each observation $o_t$ during each time step, each data point corresponding to an observed point. Pixels in $o_t$ are inversely projected into 3D coordinates in the scene using the depth channel, the camera intrinsics, as well as agent's coordinates and heading that are assumed to be available, as in~\cite{DBLP:conf/nips/WaniPJCS20}. 
The query vector $\mathbf{q}$ corresponds to a 9-dimensional vector encoding a distribution over object classes ($8$ target objects and the ``\textit{background}'' class). Let us recall that while the training of the representation is supervised, this supervision cannot use ``ground-truth'' information available only during training. All supervision information is required to be predicted from the data available to the agent even after deployment. We predict object class information through a semantic segmentation model $p$, which is applied to each current RGB-D observation $o_t \in \mathbb{R}^{h \times w \times 4}$, recovering the output segmentation map $m_t \in \mathbb{R}^{k \times l \times 9}$.
The model has been pre-trained on the segmentation of the different target objects, i.e. coloured cylinders, and is not fine-tuned during training of the agent itself.

Training data pairs $(x_i^*, q_i)$ are sampled from this output. The supervised coordinates $x_i^*$ are simply the mean 3D coordinates of each feature map cell, after inverse projection. The query vector $\mathbf{q}_i$ is the distribution over semantic classes. After the replay buffer is updated, a training batch must be sampled to update the neural field. One fourth of the samples in the batch of size $b$ correspond to the $b/4$ last steps. The rest are sampled from the previous steps in the replay buffer. Uniform sampling is also performed among pairs collected at a given time step.

\paragraph{Estimating uncertainty} --- is an essential component, as querying yet unseen objects will lead to wrong predictions, which the agent needs to recognize as such, and discard. The estimation of uncertainty in neural networks is an open problem, which has been previously addressed through different means, including drop out as a Bayesian approximation \cite{GalDropout2014}, variational information bottlenecks \cite{shah_rapid_2021}, density estimation \cite{NormFlowsPAMI2021}, and others. In this work, we approximate a density estimate in the scene replay buffer by calculating the minimum Euclidean distance between the input query and all embeddings in the  replay buffer at the current time step. The method is simple and efficient and does not require explicitly fitting a model to estimate the marginal distribution $p(\mathbf{q})$, in particular as the uncertainty representation is latent, can be un-normalized as not required to be a probability.

\subsection{Occupancy and Exploration Implicit Representation $ f_o$}
\label{sec:occupancyrepresentation}
\noindent
Unlike $f_s$, the occupancy representation $f_o$ is closer to classical implicit representations in robotics, e.g. \cite{sucar2021imap,ortiz_isdf_2022,li_learning_2022,adamkiewicz2022vision}, which map spatial coordinates to variables encoding information on navigable area like occupancy or signed distances. Different to previous work, our representation also includes exploration information, which changes over time during the episode. Once explored, a position changes its class, which makes our neural field \textit{dynamic}. Another difference with $f_s$ is that the latter deals with 3D coordinates while $f_o$ is a topdown 2D representation.
Inspired by \cite{xie2021neural,tancik2020fourier}, the model uses Fourier features $\phi$ extracted from the 2D coordinates $\mathbf{x}$ previously normalized $\in [0, 1]$,
\begin{equation}
\phi = (\cos(\mathbf{x}{\*}2^0), \sin(\mathbf{x}{\*}2^0), ..., \cos(\mathbf{x}{\*}2^{\frac{p}{4}}), \sin(\mathbf{x}{\*}2^{\frac{p}{4}})). \\
\end{equation}
The network $f_o$ is a 3-layer MLP with ReLu intermediate activations and a softmax function at the output layer. Hidden layers have $512$ neurons, and $p=40$.

\myparagraph{Mapping/Training} The implicit representation is updated minimizing the Cross Entropy loss between the prediction $\mathbf{s}$ of the neural field and the supervised label $\mathbf{s^*}$ of three classes \textit{\{Obstacle, Navigable, Unexplored}\}, as  
$
\mathcal{L}_o = - \sum_{c=1}^{3} \mathbf{s_c^*} \log  \mathbf{s}_c
$. As for the \textit{Semantic Finder}, training data pairs $(\mathbf{s}^*,\mathbf{s})$ are created through inverse perspective projection of the pixels of the observation $o_t$ into 3D scene coordinates. Thresholding the $z$ (height) coordinate decides between \textit{Navigable} and \textit{Obstacle} classes. Points with a $z$ coordinate higher than a certain threshold are discarded. The replay buffer is balanced between both classes, and only samples of the last 1000 steps are kept. Samples of the \textit{Unexplored} class are not stored.

The replay buffer is sampled similarly to the one for the \textit{Semantic Finder}. However, additional samples for the \textit{Unexplored} class are created by sampling uniformly inside the scene, for speed reasons simply ignoring conflicts with explored areas and treating them as noisy labels.

\subsection{Global Occupancy Read $r$ --- handling reparametrization invariance}
\label{sec:globalread}
\noindent
The global Occupancy reader $r$ allows to query the occupancy information of the scene globally, beyond point-wise information, and as such is a trainable mapping from the space of functions $f_o(.;.)$ to an embedding space $\mathbf{e}$. In particular, two functions $f_o$ and $f'_o$ s.t. $f_o(\mathbf{x}) = f'_o (\mathbf{x}) \ \forall \mathbf{x}$ should be mapped to identical or close embeddings. However, as the occupancy networks $f_o$ are implemented as MLPs, any given instance $f_o(.;\theta_o)$ parameterized by trainable weights $\theta_o$ can be reparametrized by any permutation of hidden units, which leads to permutations of the rows and columns, respectively, of two weight matrices, its own and the one of the preceding layer. This reparameterization keeps the represented functions identical, although their representations as weight vectors are different.

To favor learning a global occupancy reader which is invariant w.r.t these transformations, we implement it as a transformer model with self-attention \cite{vaswani2017attention} --- this, however, does not enforce full invariance. The model takes as input a sequence of tokens $(w_1, ..., w_N)$, where $w_i \in \mathbb{R}^a$ is a learned linear embedding of the incoming weights of one neuron within the implicit representation $f_o$, and $N$ is the number of neurons of $f_o$. Each token is summed with a positional encoding in the form of Fourier features. An additional ``\textit{CLS}'' token with learned embedding is concatenated to the input sequence. The reader is composed of 4 self-attention layers, with 8 attention heads. The output representation of the ``\textit{CLS}'' token is used as the global embedding of the implicit representation.

\myparagraph{Training}
The global reader $r$ is trained with full supervision from a dataset of $25k$ trajectories composed of MLP weights $\theta_{o,i}$ and absolute maps $\mathbf{M}_i, \ i..1..25k$. Each map is a metric tensor providing occupancy information extracted from the corresponding implicit representation, i.e. $\mathbf{M}_i(\mathbf{x}_y, \mathbf{x}_x) = f_o(\mathbf{x}, \theta_{o,i})$. The dataset also contains an ego-centric version $\mathbf{M}'_i$ of each map, which is centered on the agent and oriented depending on its current heading. The reader $r$ is trained in an Encoder-Decoder fashion, where $r$ plays the role of the encoder,
\begin{equation}
\textstyle{
    \mathbf{e}_i = r(\theta_{o,i}), \quad \hat{\mathbf{M}}_i = Dec(e_i, p_i),
}
\end{equation}
where $p_i$ is the agent pose (position and heading), necessary to decode ego-centric information.
We minimize a cross entropy loss on the prediction of ego-centric maps, 
\begin{equation}
\textstyle{
\mathcal{L}_g = - \sum_i \sum_k \sum_l \sum_{c=1}^{3} \mathbf{M}'^*_{i,c}(k,l) \log  \mathbf{M}'_{i,c}(k,l)
}
\end{equation}

Directly training this prediction proved to be difficult. We propose a procedure involving several steps, detailed in section~\ref{sec:training_global_reader}. After the training phase, the reader $g$ is used in the perception + mapping module of the agent as given in equation (\ref{eq:perception}), and kept frozen during agent RL training.

\subsection{Training the Agent} 
\label{sec:trainingagent}
\noindent
The agent is trained with RL, more precisely Proximal Policy Optimization (PPO)~\cite{schulman2017proximal}. The inner training loops of the implicit representations are supervised ({\color{RedArrowChris} \textbf{red arrows}} in Figure~\ref{fig:overview}) and occur at each time step in the forward pass, whereas the RL-based outer training loop of the agent occur after $N$ acting steps (\textbf{black arrows} in Figure~\ref{fig:overview}).
As the perception module used to generate training data from RGBD-observations for the \textit{Semantic Finder} $f_s$ is independent of the visual encoder $c$ in the agent (see Figure \ref{fig:overview}), and as its query $\mathbf{q}_t$ is fixed to the navigation goal $\mathbf{g}_t$, there is no need to track the weights $\theta_s$ at each time step in order to backpropagate the PPO loss (outer training). This is a key design choice of our method. 

\myparagraph{Training assumptions} 
We do not rely on the existence of global GT maps for occupancy, as $f_o()$ and $r()$ were trained on observation data from agent trajectories only. However, similar to \cite{marza2021teaching,EpisodicTransformersICCV2021}, we exploit object positions in simulation, during training only; Moreover, we  require pixel-wise segmentation masks during training. We believe that this does not change requirements, as the goal is to fully exploit 3D photo-realistic simulators as a data source and see how far the field can go with this. Generalization requirements are unchanged: we require our agent to be able to generalize to new unseen scenes. Generalization to unseen object categories is not targeted, and in \textit{MultiON} task setup not possible for any agent, as object positions and labels are required for reward calculation.

\subsection{Training the global reader}
\label{sec:training_global_reader}
\begin{figure*}[t]
    \centering
    \includegraphics[width=0.9\textwidth]{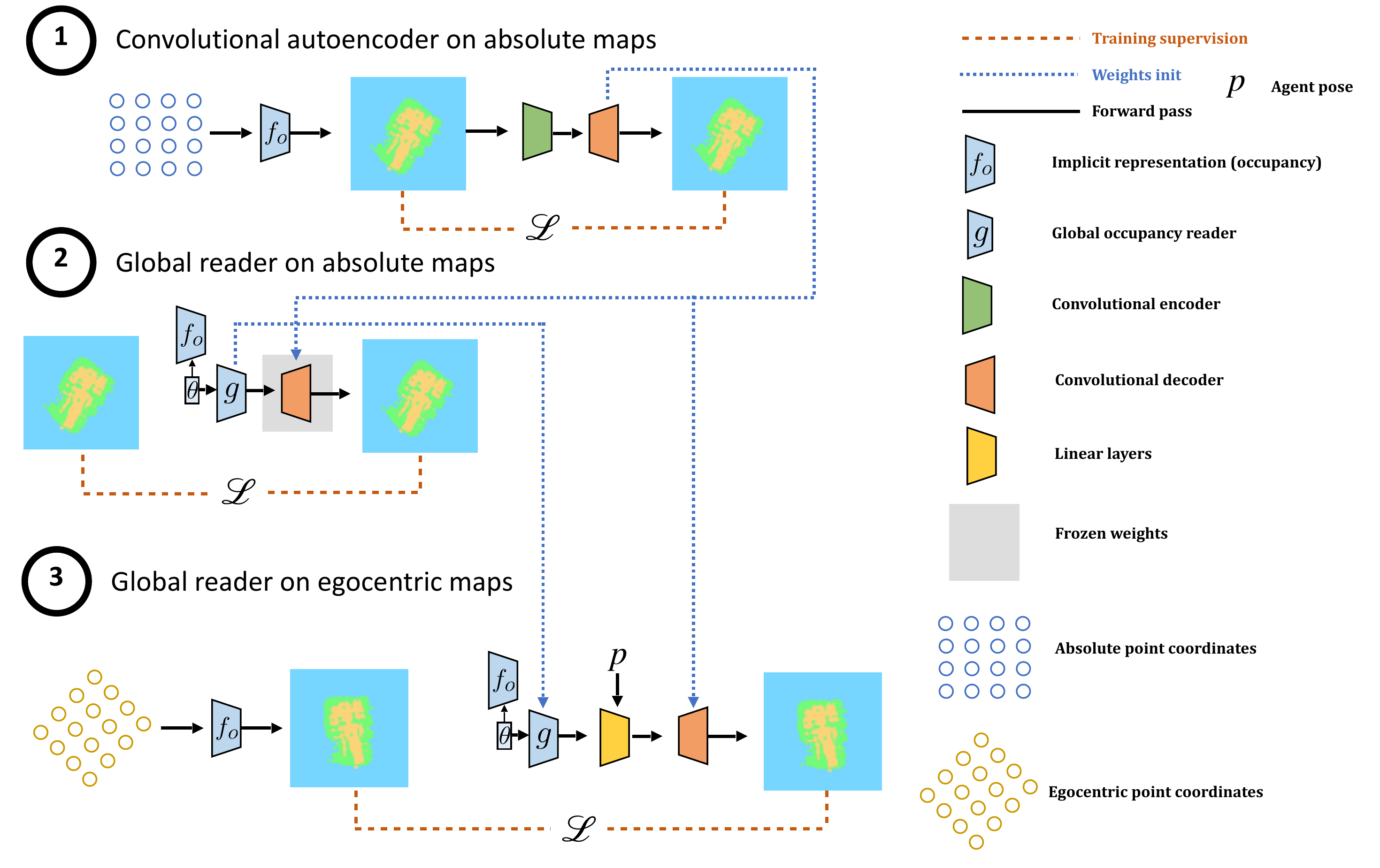}
    \caption{\textbf{Training the Global Reader} in $3$ steps: \ding{192} training a convolutional auto-encoder on absolute maps; \ding{193} extension to inputs equal to weights $\theta_o$ = the gobal reader; \ding{194} extention to predicting ego-centric maps, better suited to navigation problems.}
    \label{fig:global_reader_train}
\end{figure*}

\noindent
The proposed training procedure for the global reader $r$ (performed \textit{before} training the agent) can be split into $3$ phases. The architecture for the convolutional decoder $Dec()$ is kept the same in all of them. The hyperparameters of its different layers are detailed in the Supplementary Material. It is composed of $6$ transpose convolution layers along with batch norm layers and ReLU activations, except for the last layer with a softmax activation. Figure~
\ref{fig:global_reader_train} provides an overview of the $3$ steps involved in the training of the global reader.

\begin{description}

\item[Fully convolutional autoencoder] --- 
The first step is to train a fully convolutional autoencoder on the set of absolute maps $\mathbf{M}_i, \ i..1..25k$. Only the decoder weights are kept.

\item[Global reader training on absolute maps] ---
The second step consists in training the global reader to output embeddings fed to the frozen decoder from the previous step. The objective is to reconstruct absolute maps from the weights of the implicit representation. The global reader weights are kept after this training phase.

\item[Global reader finetuning on egocentric maps] ---
The global reader, whose weights are initialized from the weights obtained in the previous step, is now trained along with the same decoder from the first step (also used in the second step) on the set of egocentric maps $\mathbf{M}'_i, \ i..1..25k$. Both global reader and decoder are finetuned. The output of the global reader is not directly fed to the decoder, but is passed through linear layers in order to fuse information about first the position of the agent, and then its heading because this time the right operations of shift and rotation must be applied in order to reconstruct egocentric maps.

\item[Integrating the pre-trained global reader into the agent] ---
After this pre-training in $3$ steps, the weights of the global reader are frozen and not updated during the RL training. However, a linear layer is learnt to project the $576$-dim embedding from the global reader into a $256$-dim representation fed to the GRU. This linear layer is trained from reward signals.

\end{description}

\begin{table*}  \centering
  \caption{\label{table:ablation}\textbf{Impact of the implicit representations:} navigation performance on \textit{MultiON val} and \textit{MultiON test}. S=$f_s$ activated, O=$f_o$ activated in the corresponding training period (see text). Top/$\mu$: means over 3 runs; Bottom/$\uparrow$: best validation seeds over 3 runs.}  
  \centering
  {\small
\setlength{\aboverulesep}{0pt}
\setlength{\belowrulesep}{0pt}
  \begin{tabular}{c cc cc cc cccc c cccc}
    \toprule
    & \multicolumn{2}{c}{$0-30$} & \multicolumn{2}{c}{$30-50$} & \multicolumn{2}{c}{$50-70$} & \multicolumn{4}{c}{--- Val ---} &&
    \multicolumn{4}{c}{--- Test ---} \\
    & S & O & S & O & S & O &
    \textbf{Success} & \textbf{Progress} & \textbf{SPL} & \textbf{PPL}  && 
    \textbf{Success} & \textbf{Progress} & \textbf{SPL} & \textbf{PPL}  \\
    \midrule
    
    \multirow{3}{*}{$\mu$} & $-$ & $-$ & $-$ & $-$ & $-$ & $-$ & {\scriptsize $33.2 \pm$} {\tiny 1.2} & {\scriptsize $ 49.0 \pm$} {\tiny 1.1} & {\scriptsize $21.2 \pm$} {\tiny 0.5} & {\scriptsize $31.6 \pm$} {\tiny 1.2} && {\scriptsize $42.3 \pm$}  {\tiny 1.5}& {\scriptsize $56.7 \pm$}  {\tiny 0.9} & {\scriptsize $28.1 \pm$}  {\tiny 1.0} & {\scriptsize $37.8 \pm$}  {\tiny 1.8} \\
    
    & $-$ & $-$ & $\checkmark$ & $-$ & $\checkmark$ & $-$ & {\scriptsize $37.8 \pm$} {\tiny 1.6} & {\scriptsize $52.3 \pm$} {\tiny 0.9} & {\scriptsize $26.35 \pm$} {\tiny 1.5} & {\scriptsize $36.5 \pm$} {\tiny 0.8} && {\scriptsize $\textbf{47.0} \pm$} {\tiny 1.7} & {\scriptsize $\textbf{60.5} \pm$} {\tiny 1.6} & {\scriptsize $34.5 \pm$} {\tiny 0.8} & {\scriptsize $44.2 \pm$} {\tiny 1.0} \\
    
    & $-$ & $-$ & $\checkmark$ & $-$ & $\checkmark$ & $\checkmark$ & {\scriptsize $\textbf{38.5}\pm$} {\tiny 4.6} & {\scriptsize $\textbf{52.5} \pm$} {\tiny 4.8} & {\scriptsize $\textbf{28.2} \pm$} {\tiny 2.1} & {\scriptsize $\textbf{38.3} \pm$} {\tiny 1.7} && {\scriptsize 46.7 $\pm$} {\tiny 3.0} & {\scriptsize 60.1 $\pm$} {\tiny 3.1} & {\scriptsize \textbf{35.1} $\pm$} {\tiny 1.4} & {\scriptsize \textbf{44.8} $\pm$} {\tiny 1.0}  \\
\arrayrulecolor{MidRuleGray}
\specialrule{.1pt}{0.1pt}{0.1pt} 

\multirow{3}{*}{$\uparrow$} & $-$ & $-$ & $-$ & $-$ & $-$ & $-$ & {\small $32.1$} & {\small $47.7$}  & {\small $21.6$} & {\small $32.6$}  && {\small $41.0$}  & {\small $55.9$}   & {\small $28.9$}   & {\small $39.0$} \\

& $-$ & $-$ & $\checkmark$ & $-$ & $\checkmark$ & $-$ & {\small $38.1$} & {\small $51.9$}  & {\small $27.3$} & {\small $37.1$}  && {\small $48.6$}  & {\small $61.5$}   & {\small $35.2$}   & {\small $44.2$} \\

& $-$ & $-$ & $\checkmark$ & $-$ & $\checkmark$ & $\checkmark$ & {\small $\textbf{43.1}$} & {\small $\textbf{56.8}$}  & {\small $\textbf{30.5}$} & {\small $\textbf{40.1}$}  && {\small $\textbf{49.7}$}  & {\small $\textbf{63.4}$}   & {\small $\textbf{36.4}$}   & {\small $\textbf{45.9}$} \\

\arrayrulecolor{black}    
    \bottomrule
  \end{tabular}
  }
\end{table*}

\begin{table*}
\setlength{\aboverulesep}{0pt}
\setlength{\belowrulesep}{0pt}
  \caption{\label{table:sota}\textbf{Comparison with SOTA methods} on \textit{MultiON test}. $\dagger$=performance taken from~\cite{marza2021teaching}. ``\textit{AUX}'' = auxiliary losses using privileged information~\cite{marza2021teaching}. ``\textit{ORC}''=non-comparable, uses oracle information. $\rho$ = pre-training of input encoders from~\cite{marza2021teaching}. $\alpha$ = finetuning of input encoders with RL. $\gamma$ = implicit representations are accessible to the agent since the beginning of  RL training (\textit{w/o curriculum}).}
{\small
  \centering
  \begin{tabular}{l c c c c c c c c c c}
\toprule
& Agent & \textbf{$\rho$} & \textbf{$\alpha$} & \textbf{$\gamma$} & \textbf{Success} & \textbf{Progress} & \textbf{SPL} & \textbf{PPL} & AUX & ORC\\
\midrule
(a) & OracleMap$^\dagger$~\cite{DBLP:conf/nips/WaniPJCS20} & ${-}$ & ${\checkmark}$ && $50.4 \pm$ {\tiny 3.5} & $60.5 \pm$ {\tiny 3.1} & $40.7 \pm$ {\tiny 2.2} & $48.8 \pm$ {\tiny 1.9} & ${-}$ & ${\checkmark}$\\

(b) & OracleEgoMap$^\dagger$~\cite{DBLP:conf/nips/WaniPJCS20} & ${-}$ & ${\checkmark}$ && $32.8 \pm$ {\tiny 5.2} & $47.7 \pm$ {\tiny 5.2} & $26.1 \pm$ {\tiny 4.5} & $37.6 \pm$ {\tiny 4.7} & ${-}$ & ${\checkmark}$\\
\arrayrulecolor{MidRuleGray}
\specialrule{.1pt}{0.1pt}{0.1pt}
(c) & NoMap$^\dagger$~\cite{DBLP:conf/nips/WaniPJCS20} & ${-}$ & ${\checkmark}$ && $16.7 \pm$ {\tiny 3.6} & $33.7 \pm$ {\tiny 3.3} & $13.1 \pm$ {\tiny 2.4} & $26.0 \pm$ {\tiny 1.7} & $-$ & ${-}$\\

(d) & ProjNMap$^\dagger$~\cite{Henriques_2018_CVPR} & ${-}$ & ${\checkmark}$ && $25.9 \pm$ {\tiny 1.1} & $43.4 \pm$ {\tiny 1.0} & $18.3 \pm$ {\tiny 0.6} & $30.9 \pm$ {\tiny 0.7} & ${-}$ & ${-}$\\

\specialrule{.1pt}{0.1pt}{0.1pt}

(e) & NoMap & ${\checkmark}$ & ${-}$ && $42.3 \pm$ {\tiny 1.5} & $56.7 \pm$ {\tiny 0.9} & $28.1 \pm$ {\tiny 1.0} & $37.8 \pm$ {\tiny 1.8} & $-$ & ${-}$\\

(f) & ProjNMap~\cite{Henriques_2018_CVPR} & ${\checkmark}$ & ${-}$ && $39.7 \pm$ {\tiny 2.3} & $55.4 \pm$ {\tiny 1.4} & $28.7 \pm$ {\tiny 1.1} & $40.1 \pm$ {\tiny 1.9} & ${-}$ & ${-}$\\

(g) & Implicit (Ours) \textit{w/ curriculum} \textit{w/ pre-train} & ${\checkmark}$ & ${-}$ & ${-}$ &  $46.7 \pm$ {\tiny 3.0} & $60.1 \pm$ {\tiny 3.1} &  $35.1 \pm$ {\tiny 1.4} &  $44.8 \pm$ {\tiny 1.0} & ${-}$ & ${-}$\\

\specialrule{.1pt}{0.1pt}{0.1pt}

(h) & ProjNMap + \textit{AUX} ~\cite{marza2021teaching}& {\footnotesize N/A} & ${\checkmark}$ & {\footnotesize N/A} & 57.7 $\pm$ {\tiny 3.7} & 70.2 $\pm$ {\tiny  2.7} & 37.5 $\pm$ {\tiny 2.0} & 45.9 $\pm$ {\tiny 1.9} & ${\checkmark}$ & ${-}$\\

(i) & Implicit (Ours) \textit{w/o curriculum} \textit{w/ pre-train} + \textit{AUX} & ${\checkmark}$ & ${\checkmark}$ & ${\checkmark}$ &  $58.3 \pm$ {\tiny 0.8} & $69.4 \pm$ {\tiny 1.1} &  $\textbf{43.8} \pm$ {\tiny 1.0} &  $\textbf{52.1} \pm$ {\tiny 1.6} & ${\checkmark}$ & ${-}$\\

(j) & Implicit (Ours) \textit{w/o curriculum} \textit{w/o pre-train} &  ${-}$ & ${\checkmark}$ & ${\checkmark}$ & $54.8 \pm$ {\tiny 3.6} & $68.0 \pm$ {\tiny 3.4} &  $41.7 \pm$ {\tiny 1.9} &  $51.3 \pm$ {\tiny 1.6} & ${-}$ & ${-}$ \\

(k) & Implicit (Ours) \textit{w/o curriculum} \textit{w/o pre-train} + \textit{AUX} &  ${-}$ & ${\checkmark}$ & ${\checkmark}$ & $57.9 \pm$ {\tiny 2.0} & $69.5 \pm$ {\tiny 0.6} &  $\textbf{43.3} \pm$ {\tiny 2.2} &  $\textbf{51.9} \pm$ {\tiny 3.7} & ${\checkmark}$ & ${-}$ \\

\arrayrulecolor{black}
\bottomrule
\end{tabular}
}
\end{table*}

\begin{table}
\setlength{\aboverulesep}{0pt}
\setlength{\belowrulesep}{0pt}
\centering
{\small 
  \begin{tabular}{c c c c c c c c}
\toprule
Uncertainty & \textbf{Success} & \textbf{Progress} & \textbf{SPL} & \textbf{PPL}\\
\midrule
 ${-}$ & $35.4 \pm$ {\tiny 3.0} & $49.7 \pm$ {\tiny 3.3} & $29.4 \pm$ {\tiny 2.0} & $40.9 \pm$ {\tiny 2.4}\\

 ${\checkmark}$ & $\textbf{43.4} \pm$ {\tiny 3.1} & $\textbf{58.0} \pm$ {\tiny 3.0} & $\textbf{35.1} \pm$ {\tiny 0.8} & $\textbf{46.4} \pm$ {\tiny 1.0}\\ 
\arrayrulecolor{black}
\bottomrule
\end{tabular}
}
\caption{\label{table:uncertainty_ablation} \textbf{Uncertainty:} comparing training w/ semantic input only, no occupancy, from the beg. of training, w/ and w/o uncertainty.\vspace*{-5mm}}  
\end{table}

\section{Experiments}

\myparagraph{\textit{MultiON} task}  we target the 3-ON version of the \textit{MultiON} task~\cite{DBLP:conf/nips/WaniPJCS20}, where the agent deals with sequences of $3$ objects, each belonging to one of $8$ classes (cylinders of different colors). At each step, the observation $\mathbf{o}_t$ is an RGB-D image of size $256{\times}256{\times}4$ and the target class is a one-in-K ($K{=}8$) vector. The action space is discrete: \{\textit{Move forward} $0.25m$, \textit{Turn left} $30\degree$, \textit{Turn right} $30\degree$, \textit{Found}\}. An episode is considered successful if the agent finds all of the targets before the time limit ($2,500$ environment steps), and chooses the \textit{Found} action for each one at a distance closer to $1.5m$. Calling \textit{Found} incorrectly terminates episode as a failure. An access to perfect odometry information (localization and heading) was assumed in~\cite{DBLP:conf/nips/WaniPJCS20} as the standard protocol.

\myparagraph{Dataset and metrics} The agent is trained on the Matterport3d~\cite{chang2018matterport3d} dataset. We followed the standard train/val/test split over scenes (denoted \textit{MultiON train}, \textit{MultiON val}, \textit{MultiON test}): $61$ training, $11$ validation and $18$ test scenes. The train, val and test splits are respectively composed of $50,000$, $12,500$ and $12,500$ episodes per scene. Reported results on the val and test sets (Tables~\ref{table:ablation} and \ref{table:sota}) were computed on a subset of $1,000$ randomly sampled episodes.
We report standard metrics as used in the navigation literature (and in~\cite{DBLP:conf/nips/WaniPJCS20}):
\textit{Success}: percentage of successful episodes --- all objects are reached respecting order, time; \textit{Progress}: percentage of objects successfully found (respecting order, time);
\textit{SPL}: Success weighted by Path Length, extending the original SPL metric  \cite{DBLP:journals/corr/abs-1807-06757}  to \textit{MultiON};
\textit{PPL}: Progress weighted By Path Length (the official \textit{MultiON} challenge metric).

\myparagraph{\textit{Global reader} dataset} The \textit{Global reader} $r$ was trained on a dataset of $25k$ trajectories obtained from rollouts performed by a baseline agent~\cite{marza2021teaching}. $95\%$ was used for training and the rest for validation. On these trajectories we first trained the occupancy representation $f_o$ ``\textit{in-situ}'', i.e. as if it were deployed on the agent, and we recorded training samples $i$ for training the reader $r$: pairs of network weights $\theta_{o,i}$ and associated maps $M_i$ obtained by iteratively querying the implicit representation. Ego-centric maps were generated from the absolute ones and both were cropped around their center.

\myparagraph{Perception module dataset} The perception module $p$ was trained to segment the different target objects. The generated dataset is composed of $132k$ pairs of RGB-D observations and segmentation masks. Samples for $4$ scenes were kept as a validation set.

\myparagraph{Training details} We use the reward function given in~\cite{DBLP:conf/nips/WaniPJCS20} for RL/PPO training (see supplementary material), and train all agents for $70M$ steps as in~\cite{marza2021teaching}. For all agents in Table~\ref{table:ablation} and some in Table~\ref{table:sota} (\textit{w/ pre-train}: $\checkmark$ in $\rho$ column), the encoders (visual encoder $c$, as well as goal and previous action embedding layers, see Figure \ref{fig:overview}) are pre-trained with a baseline, which corresponds to the \textit{ProjNeuralMap} agent trained with auxiliary losses~\cite{marza2021teaching}. This is done to faster training, as it will be shown later (in Table~\ref{table:sota}) that the same final performance can be reached without this initial pre-training of encoders. Training and evaluation hyper-parameters, as well as architecture details have been taken from~\cite{DBLP:conf/nips/WaniPJCS20}. All reported quantitative results are obtained after $3$ training runs for each model.

\begin{figure*}  \centering
\begin{floatrow}
    \ffigbox[11cm]{%
  \includegraphics[width=11cm]{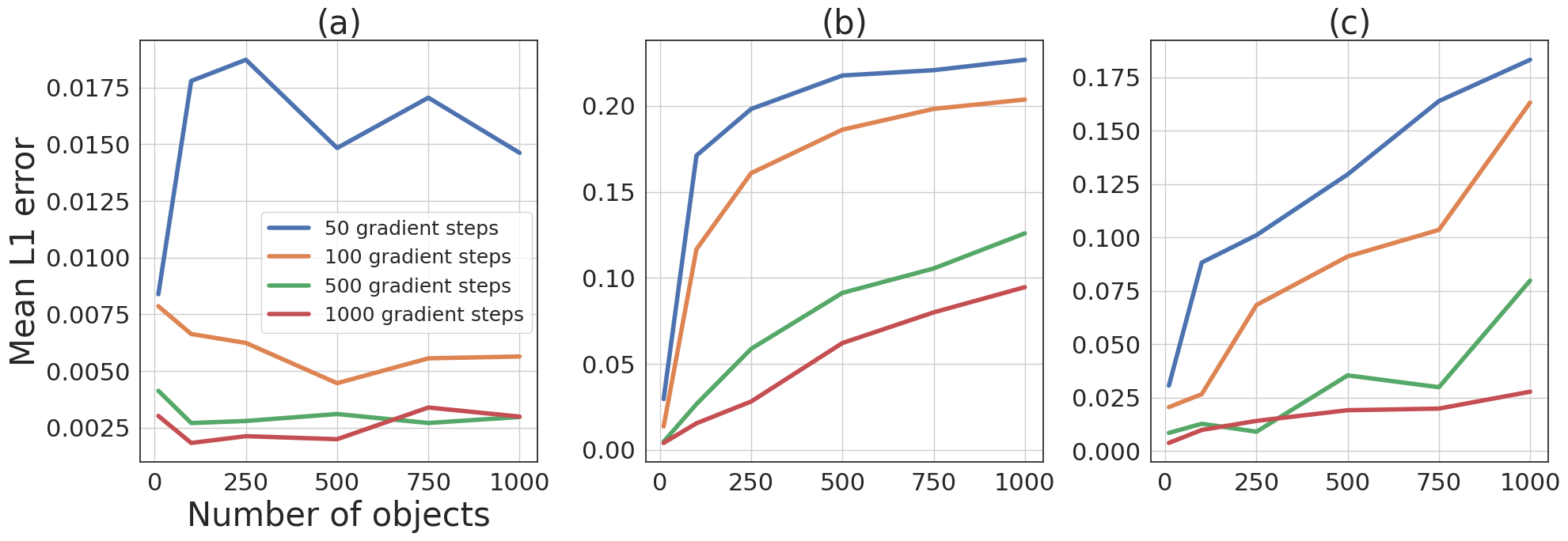}
}{%
  \caption{\label{fig:benchmark_nb_obj_dummy} \textbf{Capacity of the semantic representation:} we report mean distance prediction error (normalized $\in[0,1]$) as a function of the number of stored objects. Replay buffers are composed of dummy queries: (a) one-hot queries with same dimension as number of objects; (b) random query with dimension $9$; (c) random query with same dimension as number of objects. \vspace*{-5mm}}
}
\ffigbox[5.5cm]{
\includegraphics[width=5cm]{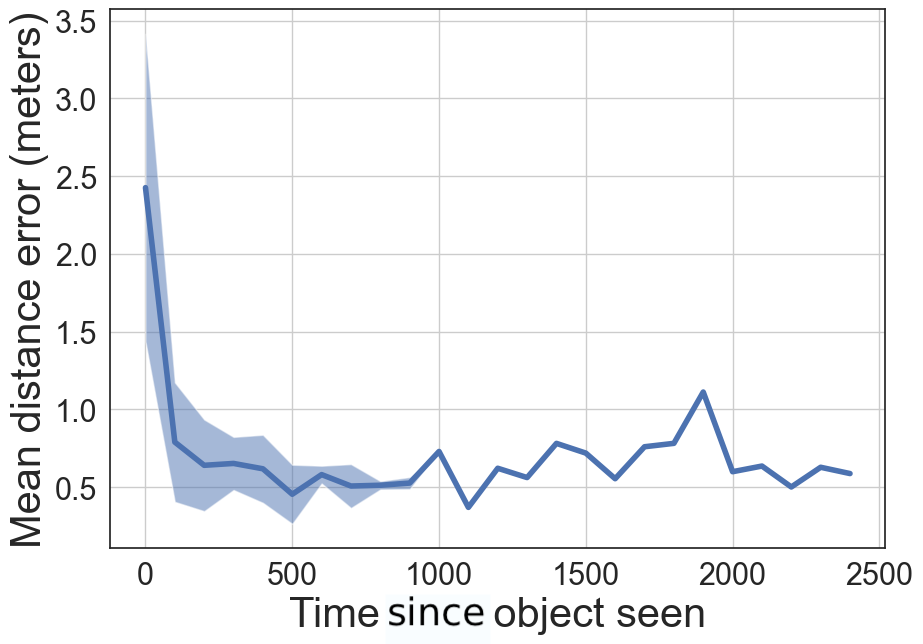}
}{
\caption{\label{fig:dist_error_semantic_net}\textbf{Lifelong learning of the semantic repr. $f_s$:} we report mean error in meters, test set, 600 episodes, as a function of the number of time steps since the object was first seen in the episode (t=0). The error falls immediately and stays low over the episode.\vspace*{-5mm}}
}
\end{floatrow}
\end{figure*}

\myparagraph{Impact of the implicit representations} Table~\ref{table:ablation} shows the impact of the two implicit representations on navigation (top: means over 3 runs; bottom: best validation seeds over 3 runs). 
To keep compute requirements limited and decrease sample complexity, in these ablations we do not train the full agent from scratch, in particular since the early stages of training are spent on learning basic interactions. We decompose training into three phases: $0{-}30M$ steps (no implicit representations, i.e. all entries to the agent related to $f_s$ and $f_o$ are set to $0$); $30M{-}50M$ steps (training includes the \textit{Semantic Finder} $f_s$) and finally $50M{-}70M$ steps (full model).
This 3-steps approach is used to train all agents in Table~\ref{table:ablation}, and will be denoted as \textit{curriculum} (See Table~\ref{table:sota}, \textit{w/ curriculum}: $-$ in $\gamma$ column).
All metrics on both val and test sets are improved, with the biggest impact provided by the \textit{Semantic Finder}, which was expected. We conjecture that mapping object positions is a more difficult task, which is less easily delegated to the vectorial GRU representation, than occupancy. We also see an impact of the occupancy representation, which not only confirms the choice of the implicit representation $f_o$ itself, but also its global read through $r(\theta_o)$. Training curves are given in the supplementary material.

\myparagraph{Uncertainty} has an impact on agent performance, as we show in the ablation in Table \ref{table:uncertainty_ablation}. Indeed, when training an agent with the semantic input since the beginning of training (\textit{w/o curriculum}) and no occupancy input (as the uncertainty is only related to semantic information), feeding the agent with the computed uncertainty about the output of the \textit{Semantic Finder} brings a boost in performance.

\myparagraph{Comparison with previous SOTA methods} is done in Table~\ref{table:sota}. The performance entries of these baselines are taken from~\cite{marza2021teaching}, which describes the winning entry of the \textit{CVPR 2021 MultiON} competition. 
Our method outperforms the different competing representations, even when they benefit from the same pre-training scheme and are thus completely comparable. NoMap with pre-training corresponds to the first row of Table~\ref{table:ablation}. The difference between (g)  and (i) is the use of the auxiliary tasks in~\cite{marza2021teaching}, but also that the implicit representations are available to the agent during the whole training period for (i), i.e. no decomposition into $3$ phases as in the ablations in Table~\ref{table:ablation} (\textit{w/o curriculum}). Moreover, compared to (g), in (i) the weights of the pre-trained encoders are finetuned. We see that the gains of our representations are complementary to the  auxiliary losses in~\cite{marza2021teaching}. (j) and (k) confirm this finding, showing that most of the gain compared with (h) comes from the implicit representations, with the auxiliary losses bringing an additional boost. (j) and (k) also show that, even though pre-training can help speed up RL training, similar test performance is achieved without it.

\myparagraph{Reconstruction performance of the Global Reader $r$}  although the task of reconstructing egocentric maps from occupancy functions $f_o$ (represented by $\theta_o$) is only used as a proxy task to train the Global reader $r$, we see it is a reliable
proxy for the quality of extracting the global latent vector $\mathbf{e}$ fed to the agent. 

\begin{wraptable}{l}{3.7cm}  \centering
\small{
  \begin{tabular}{c c}
    \toprule
    \textbf{Accuracy} & \textbf{Jaccard Index} \\
     $83.4$ & $56.5$ \\
    \bottomrule
  \end{tabular}
}
\caption{\label{table:mlp_reader_res}\textbf{Performance of the global reader $r$:} We report accuracy and Jaccard index.}
\end{wraptable}

\noindent
In Table~\ref{table:mlp_reader_res} we report reconstruction performance measured as accuracy and mean Jaccard Index on the validation split of the dataset used to train the reader. We judge that an accuracy of 83.4\% is surprisingly high, given that the global reader needs to reconstruct the content of the representation directly from its parameters $\theta_o$, that \textit{each implicit representation has been initialized randomly}, and that the reader is required to be invariant w.r.t. to reparameterization (see Section \ref{table:mlp_reader_res}). The task is even made harder as neural weights can be considered as an absolute representation of the env. and the reader must combine it with information about the agent pose to reconstruct an egocentric map.

\myparagraph{Capacity of the Semantic Finder} Unlike all other results presented in this paper, this experiment is performed independently of the official \textit{MultiON} benchmark. We construct a synthetic dataset to evaluate the capacity of the \textit{Semantic Finder}. More details about the generated data can be found in the supplementary material. As the granularity of the implicit representations is handled through the budget in terms of trainable parameters, we evaluate the capacity of the \textit{Semantic Finder} $f_s$ to store large numbers of objects in Figure \ref{fig:benchmark_nb_obj_dummy}: In (a), we can see that increasing the number of objects up to 1000 does not increase error with sufficient gradient updates. However, since inputs are 1-in-K, increasing the number of objects also increases capacity. In (c), we see that this does not hold for random objects (not 1-in-K), whose less structured storage requires more capacity. In (b), we keep the input dimension fixed for random objects, which further increases error.

\myparagraph{Evaluating catastrophic forgetting} we evaluate the capacity of the \textit{Semantic Finder} $f_s$ to hold the learned information over the full length of the sequence in spite of the fact that it is continuously trained. Figure~\ref{fig:dist_error_semantic_net} shows the evolution of the mean error in distance for the predicted position of queried target objects as a function of time. The error quickly goes below $1.5m$ once the object has been seen the first time ($t{=}0$ in the plot), which is distance threshold required by the \textit{MultiON} task, and stays there, providing evidence that the model does not suffer from catastrophic forgetting.

\myparagraph{Runtime performance} inspite of requiring to continously train the representations, we achieve $45$ fps during parallelized RL training, including the environment steps (simulator rendering), forward passes, representation training and RL training on a single V100 GPU. The average time of one agent forward pass, including updates of implicit representations is $20$ms on a V100 GPU, which is equivalent to $50$ fps, enough for real-time performance.

\section{Conclusion}

\noindent
We introduce two implicit representations to map semantic, occupancy and exploration information. The first estimates the position of an object of interest from a vector query, while the second encapsulates information about occupancy and explored area in the current environment. We also introduce a global read directly from the trainable weights of this  representation. Our experiments show that both implicit representations have a positive impact on the navigation performance of the agent. We also studied the scaling laws of the semantic representation and its behavior in the targeted lifelong learning problem. Future work will target differentiating through the implicit representations all the way back to the perception modules.

\textbf{Acknowledgement} ---
We thank ANR for support through AI-chair grant ``Remember'' (\small{ANR-20-CHIA-0018}).

%%%%%%%%% REFERENCES
{\small
\bibliographystyle{ieee_fullname}
\bibliography{ms}
}

\clearpage
\appendix

\begin{center}
{\Large \textbf{Appendix}}
\end{center}

\section{Training stability and curves}
\begin{figure*}[t]
    \centering
    \includegraphics[width=0.45\textwidth]{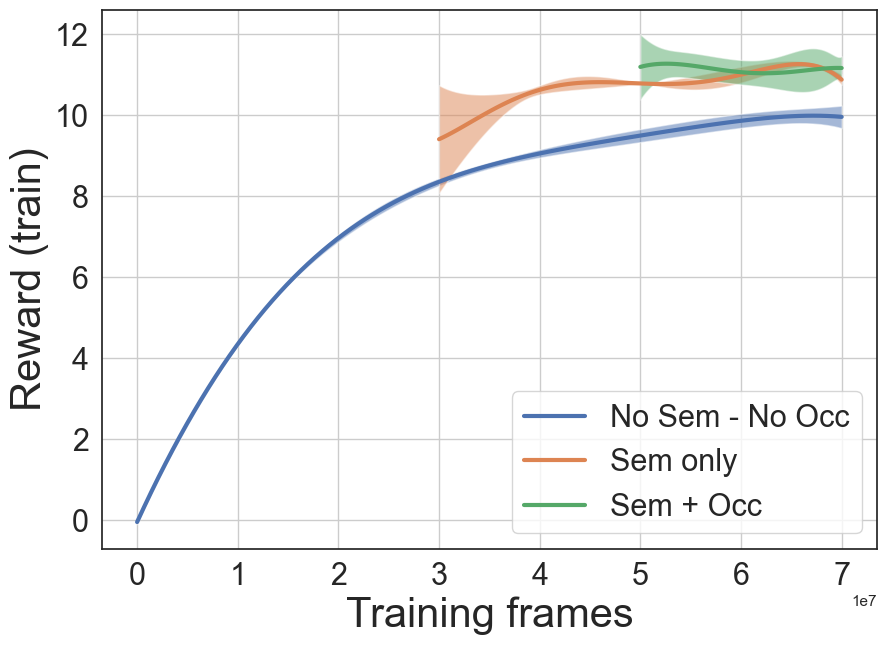}
    \includegraphics[width=0.45\textwidth]{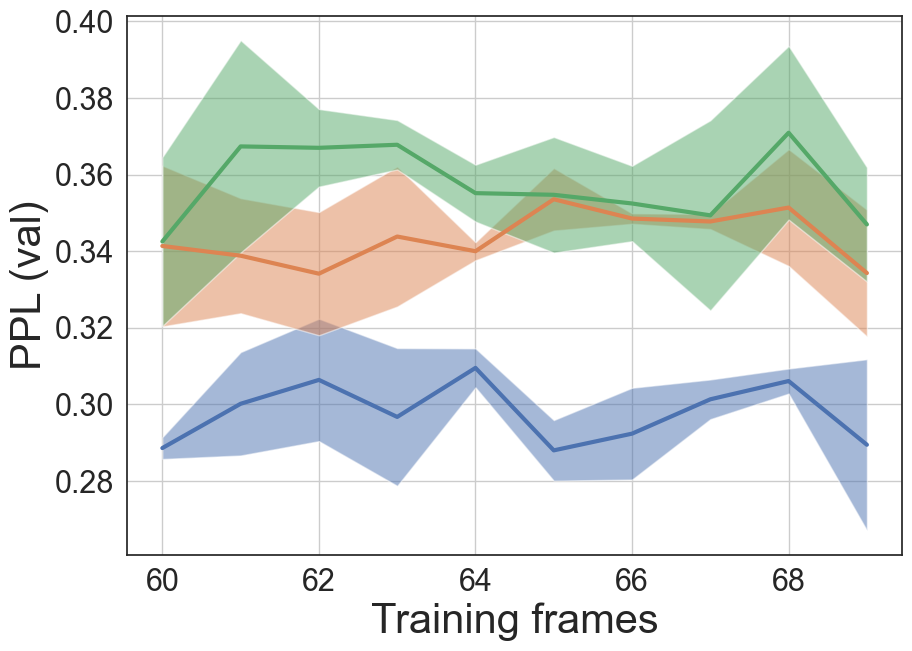}
    \hspace*{1.15cm}
    (a)
    \hspace*{7.5cm}
    (b)
    \caption{\textbf{Training stability:} (a) Evolution of the collected reward on training episodes for the $3$ models presented in Table $1$ in the main paper. (b) Evolution of PPL, the official ranking metrics in the \textit{MultiON} Challenge leaderboard, on val episodes for model checkpoints from the last $10$M training frames.}
    \label{fig:training_curves}
\end{figure*}

\noindent
Figure~\ref{fig:training_curves} shows the training curves of the $3$ different agents we compared in Table $1$ of the main paper: the recurrent baseline agent (blue), with the \textit{Semantic Finder} (orange) and with both implicit representations (green). Left, (a), we see the evolution of the training reward as a mean and standard deviation over 3 runs. Right, (b) shows PPL, the main metric chosen for ranking the agents in the \textit{MultiON} learderboard, which we show for different checkpoints during training and evaluated on the \textbf{validation set}. As can be seen, training is quite stable over runs, and adding the two representations provides a boost in performance (as already reported in Table $1$ in the main paper). 

\section{Visualization of the agent behavior}
\noindent
\subsection{Successful episode}
\begin{figure*} [p]
    \centering
    \includegraphics[width=0.55\textwidth]{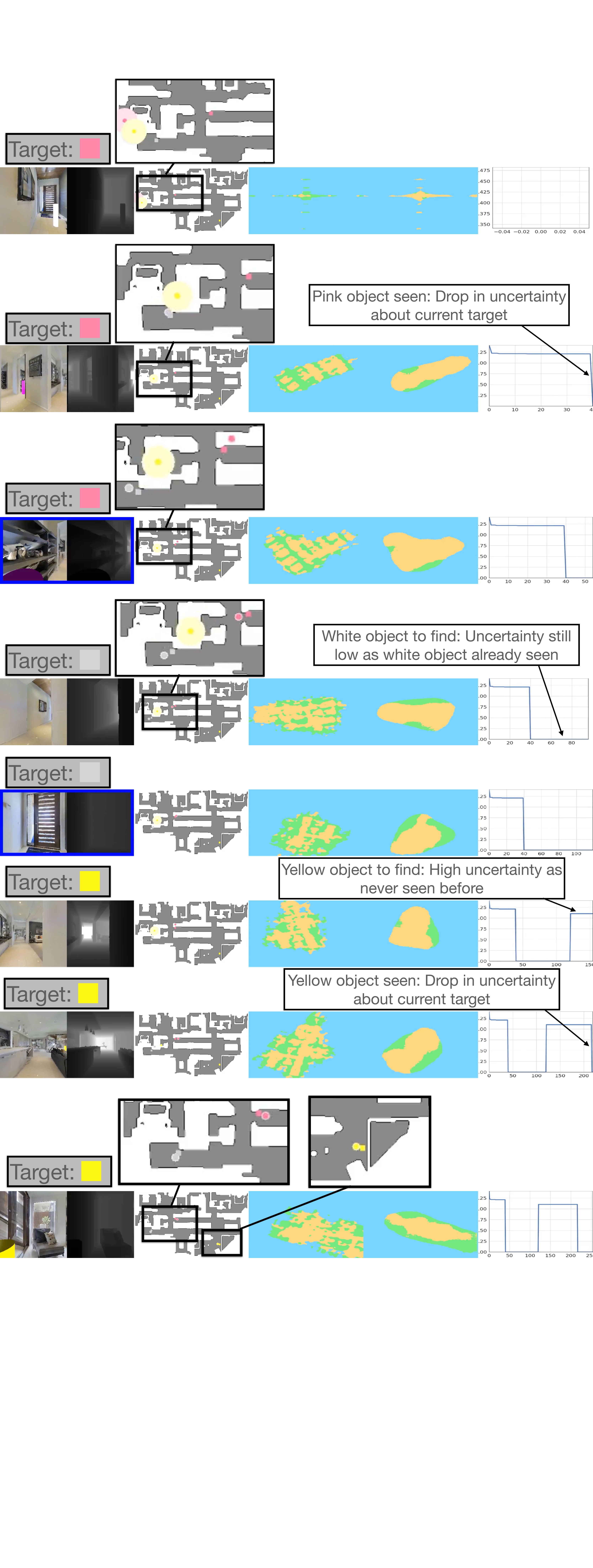}
    \caption{\label{fig:rollout} \textbf{Agent rollout on an example successful episode} from the \textit{MultiON} CVPR 2021 Challenge minival set. From left to right: RGB-D ego view, topdown map (viz only) with targets (squares) and their estimated location by the \textit{Semantic Finder} (dots, shaded region for uncertainty), map from the \textit{Occupancy and Exploration Implicit Representation}, reconstructed egomap from the global reader and CNN Decoder trained end-to-end, uncertainty of the \textit{Semantic Finder} on the currently selected target.}
\end{figure*}

\noindent
Figure~\ref{fig:rollout} provides an example for an episode rollout from the minival set of the \textit{MultiON} CVPR 2021 Challenge. The agent is equipped with both implicit representations. For each line, from left to right, we first see the RGB-D egocentric view of the agent, then  a topdown map with the three goals (white, pink and yellow squares) and their estimated location by the \textit{Semantic Finder} (white, pink and yellow dots, with a shaded region to denote uncertainty). The radius illustrating uncertainty is unit-less and only given for visualization purposes - is not available to the agent in this particular form. The third illustration shows the implicit map obtained by querying the \textit{Occupancy and Exploration Implicit Representation}, and on the right, there is the reconstructed output when feeding the embedding of the global reader to the convolutional decoder it was trained with. The last element is a curve showing the evolution of the uncertainty estimation of the \textit{Semantic Finder} on the currently provided target object.

In this episode, the agent starts with the white object within its field of view, but the first target to reach is the pink cylinder (Row $1$). As we can see, the estimation of goal positions from the \textit{Semantic Finder} are wrong, which is expected as the episode has not yet started. However, the associated uncertainty is high, allowing the agent to discard this information. The agent then explores until it observes the pink object (Row $2$). At that point, the uncertainty about the object to find drops. The estimate of the position of the pink object will be updated as training samples will be added to the semantic replay buffer. Also note that at that point the estimate of the position of the white object from the \textit{Semantic Finder} is accurate as the object has already been seen previously. The agent then goes towards the pink target object and calls the \textit{found} action (Row $3$). Estimation of the positions of pink and white objects are accurate. The next target to find is the white object. The uncertainty about the current target is low as the white object has already been observed. The agent backtracks (Row $4$) and goes towards the white object to call the \textit{found} action (Row $5$). The next goal is the yellow cylinder. At that point, the uncertainty about the current target increases as the yellow cylinder has not yet been within the agent's field of view. The agent explores (Row $6$) and when the target is within its field of view (Row $7$) the uncertainty related to the target to find drops. The agent goes towards the yellow object and calls the \textit{found} action (Row $8$). At the end of the episode, the \textit{Semantic Finder} is able to localize the $3$ objects, and the associated uncertainties are low. All objects have been successfully found, so this episode is considered as a success.

\subsection{Failure case}
\begin{figure*} [p]
    \centering
    \includegraphics[width=0.45\textwidth]{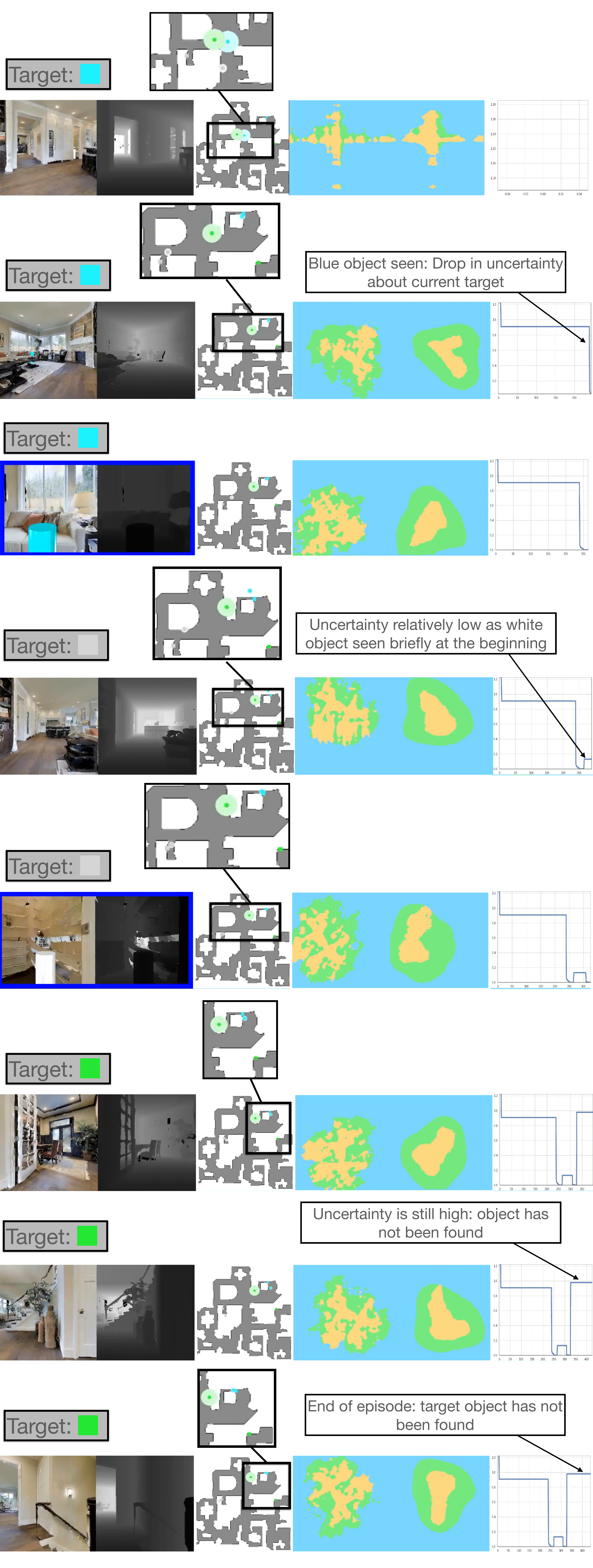}
    \caption{\label{fig:rollout_failure} \textbf{Agent rollout on an example unsuccessful episode} from the \textit{MultiON} CVPR 2021 Challenge minival set. From left to right: RGB-D ego view, topdown map (viz only) with targets (squares) and their estimated location by the \textit{Semantic Finder} (dots, shaded region for uncertainty), map from the \textit{Occupancy and Exploration Implicit Representation}, reconstructed egomap from the global reader and CNN Decoder trained end-to-end, uncertainty of the \textit{Semantic Finder} on the currently selected target.}
\end{figure*}

\noindent
Figure~\ref{fig:rollout_failure} shows an example of unsuccessful episode. The agent and the setup are the same as described in the previous subsection.
The agent can see the white target at the beginning of the episode, but it is far and largely occluded.The first target to find is the blue cylinder. It thus explores the scene until seeing the object. The uncertainty thus drops, the prediction of the target position on the map is now correct. It is interesting to see that the prediction of the location of the white target is also pretty accurate even though it was hard to detect at the beginning of the episode. The agent reaches the blue target and calls the \textit{Found} action. The next target to find is the white object. The uncertainty is not $0$ but relatively small as the agent has already seen the white target before. It goes towards the object, and when it is within its field of view, the uncertainty drops to $0$ and the location prediction on the map is accurate. The agent calls the \textit{Found} action. It has thus succeeded in finding the first two objects that were quite close to its initial position. However, as shown in the last $3$ rows, the agent does not succeed in exploring the environment enough and never finds the last target which is the green cylinder. After a few steps, it calls the \textit{Found} action at the wrong location.

\begin{table*}  \centering
  \caption{\textbf{Convolutional layers:} hyperparameter values in the different presented CNN architectures. $Dec$ is the CNN decoder trained with the global reader, $Enc$ is the visual encoder used in both the representation and reactive perceptions, $Seg$ is the segmentation head combined with $Enc$ in the representation perception module $c$.}
  \label{table:cnn_layers}
  \centering
  {\small
  \begin{tabular}{ccccccccc}
    \toprule
    Model & layer id & type & in channels & out channels & kernel size & stride & in padding & out padding \\
    
    \multirow{6}{*}{$Dec$} & $0$ & TransposeConv2D & $64$ & $32$ & $3$ & $2$ & $0$ & $0$ \\
    & $1$ & TransposeConv2D & $32$ & $32$ & $3$ & $2$ & $0$ & $0$ \\
    & $2$ & TransposeConv2D & $32$ & $16$ & $3$ & $2$ & $0$ & $0$ \\
    & $3$ & TransposeConv2D & $16$ & $8$ & $3$ & $2$ & $0$ & $0$ \\
    & $4$ & TransposeConv2D & $8$ & $8$ & $3$ & $2$ & $0$ & $0$ \\
    & $5$ & TransposeConv2D & $8$ & $3$ & $3$ & $2$ & $0$ & $1$ \\
    
    \midrule
    
    \multirow{3}{*}{$Enc$} & $0$ & Conv2D & $4$ & $32$ & $8$ & $4$ & $0$ & $-$ \\
     & $1$ & Conv2D & $32$ & $64$ & $4$ & $2$ & $0$ & $-$ \\
     & $2$ & Conv2D & $64$ & $32$ & $3$ & $1$ & $0$ & $-$ \\
     
     \midrule
    
    \multirow{3}{*}{$Seg$} & $0$ & Conv2D & $32$ & $32$ & $5$ & $1$ & $2$ & $-$ \\
     & $1$ & Conv2D & $32$ & $32$ & $5$ & $1$ & $2$ & $-$ \\
     & $2$ & Conv2D & $32$ & $9$ & $3$ & $1$ & $1$ & $-$ \\
      
    \bottomrule
  \end{tabular}
  }
\end{table*}

\section{Perception modules}
\noindent
Two different modules are used in this work. The first one, responsible for representation perception, extracts representations from the RGB-D observation to populate the training replay buffer of the \textit{Semantic Finder}. The second one, tackling reactive perception, encodes the observation into a vector fed to the GRU. This  representation of the observation is thus more directly used in the decision making process --- the name \textit{Reactive} is certainly not 100\% accurate, since the output of this module is still used to update agent memory, but this concerns only the hidden GRU memory and not the main implicit representations.

\begin{description}

\item[Reactive perception]
We use the encoder module in~[51] (ref from the main paper), which  encodes visual observations at each step. Table~\ref{table:cnn_layers} ($Enc$) presents the hyperparameters of the convolutional layers in this visual encoder. Is is composed of $3$ convolutional layers follower by a linear layer. ReLU activations are used. The embedding produced by this visual encoder is fed to the GRU module. In our work, the reactive perception module has been pre-trained with auxiliary losses, which corresponds to the method in~[34] (ref from the main paper). It is then frozen and not updated during  RL training.

\item[Representation perception]
The goal of the representation perception module is to extract vectors to be added to the \textit{Semantic Finder} training replay buffer. The backbone encoder is the same as the reactive perception module (see $Enc$ in Table~\ref{table:cnn_layers}) also pre-trained from the agent in~[34] (ref from the main paper). This network is augmented with a segmentation head and is fine-tuned end-to-end on the task of segmenting \textit{MultiON} target objects. Table~\ref{table:cnn_layers} ($Seg$) details the architecture of the segmentation head. It is composed of $3$ convolutional layers with ReLU activations, except for the last layer where a softmax activation is applied. After this training phase, the weights of the representation perception module are frozen and not updated during RL training.

\end{description}

\begin{algorithm} 
    \SetKwFunction{getSemBatch}{getSemBatch}
    \SetKwFunction{addSemSamples}{addSemSamples}
    
    \SetKwFunction{getOccBatch}{getOccBatch}
    \SetKwFunction{addOccSamples}{addOccSamples}
    
    \SetKwFunction{labelOccPos}{labelOccPos}
    
    \SetKwFunction{invProj}{invProj}
    \SetKwFunction{SGD}{SGD}
    \SetKwFunction{meanPooling}{meanPooling}
    \SetKwFunction{eval}{eval}
    
    \SetKwInOut{KwIn}{Input}
    \SetKwInOut{KwOut}{Output}
    
    \KwIn{Observation $o_t$, camera intrinsics $K$, goal $g_t$, replay buffers $r_s$ and $r_o$, weights $\theta_{s,t}$, $\theta_{o,t}$}
    
    \tcp{Adding training samples to the semantic replay buffer}
    $m_t = p(o_t)$ \\
    $n_t = \invProj(o_t, K)$ \\
    $k_t = \meanPooling(n_t)$ \\
    $r_s = \addSemSamples(r_s, m_t, k_t)$ \\
    \tcp{Adding training samples to the occupancy replay buffer}
    $l_t = \labelOccPos(n_t)$ \\
    $r_o = \addOccSamples(r_o, n_t, l_t)$ \\
    \tcp{Updating the Semantic Finder}
    \For{$i \leftarrow 0$ \KwTo $n_s - 1$}{
        $b_s = \getSemBatch(r_s)$ \\
        $\theta_{s,t} = \SGD(b_s, \theta_{s,t})$
    }
    \tcp{Updating the Occupancy and Exploration Implicit Representation}
    $loss = 0$ \\
    $j = 0$ \\
    \While{$loss > thresh$ and $j < n_o$}{
        $b_o = \getOccBatch(r_o)$ \\
        $loss = \eval(b_o, \theta_{o,t})$ \\
        $\theta_{o,t} = \SGD(b_o, \theta_{o,t})$ \\
        $j = j + 1$
    }
    \caption{Different steps necessary to update implicit representations at each agent step.}
    \label{algo:forward_pass}
\end{algorithm}

\section{Algorithmitic description of an agent forward pass}
\noindent
Algorithm~\ref{algo:forward_pass} gives a high-level overview of the different steps happening after receiving the current observation from the environment to take the most suitable action. 

\subsection{Lines $1-5$ --- Adding
\noindent
training samples to the semantic replay buffer}
\noindent
The segmentation map $m_t$ is obtained by passing the RGB-D observation $o_t$ through the representation perception module, i.e. a segmentation model pre-trained to segment the target objects (Line $2$). An inverse projection operation, denoted $invProj()$ is used to project pixels from $o_t$ into their 3D coordinates $n_t$ using the depth channel of $o_t$ and the known camera intrinsics $K$ (Line $3$). A meanpooling operation, denoted $meanPooling()$ is then applied to $n_t$ in order to obtain the mean 3D coordinates of all pixels in each cell of the segmentation map $m_t$ (Line $4$). Finally, pairs of softmax distribution over classes from $m_t$ and mean 3D coordinates are added to the training replay buffer of the \textit{Semantic Finder}. This is implemented as the $addSemSamples()$ in the algorithm (Line $5$).

\subsection{Lines $6-8$ --- Adding training samples to the occupancy replay buffer}
\noindent
The 3D coordinates of projected pixels in $n_t$ are compared with threshold values along their vertical $y$ coordinate to be either labelled as navigable space or obstacle. Only 3D points with their vertical coordinate below than another threshold value are kept. These comparisons are done in the $labelOccPos()$ function (Line $7$). Pairs of label and 3D coordinates are then sampled in order to keep the balance between the two classes and added to the training replay buffer of the \textit{Occupancy and Exploration Implicit Representation}. This is the $addOccSamples()$ function (Line $8$).

\subsection{Lines $9-13$ --- Updating the Semantic Finder}
\noindent
Two operations are repeated $n_s$ times. First a batch of training examples $b_s$ is sampled ($getSemBatch()$, line $11$). Then, the $SGD()$ function encapsulates the forward pass of $f_s$ on the sampled batch, the L1 loss computation, gradient computation and finally backpropagation. In this work, we fixed $n_s = 1$.

\subsection{Lines $14-22$ --- Updating the Occupancy and Exploration Implicit Representation}
\noindent
The implicit representation is iteratively updated for a maximum of $n_o$ steps while the error of the model ($loss$, initialized to $0$ in line $15$) is higher than a threshold. Same as for the \textit{Semantic Finder}, a training batch $b_o$ is first sampled ($getOccBatch()$, line $18$). The model is then evaluated on samples from $b_o$ (Line $19$). The $SGD()$ function is then applied to update the implicit representation. In this work, we chose $n_o = 20$ and $thresh = 0.3$.

\section{Capacity of the Semantic Finder} \noindent
This section provides further details about the study conducted to evaluate the impact of number of objects and the nature of their representation on the capacity of the Semantic Finder to memorize their position, and is complementary to the paragraph ``\textit{Capacity of the Semantic Finder}'' in Section 4 and Figure 3 of the main paper. 

To be flexible in the amount of objects we can use, we perform these experiments independently of the official \textit{MultiON} benchmark. We consider three new scenarios, and for each one, a dataset is generated and used to train the Semantic Finder. All datasets are made of (query, position) pairs with positions being uniformly sampled between arbitrary scene bounds (between $0$ and $1$ along each axis). For each dataset, we also create variants varying the sample size. To reduce the amount of hyper-parameters (e.g. batch size), we ressort to gradient descent as opposed to stochastic gradient descent, i.e. each gradient step is computed over the whole dataset. The considered metrics is the mean $L_1$ error on the prediction of positions as a function of the number of objects to memorize. The difference between the three scenarios is in the nature of queries associated with positions, and each scenario corresponds to a sub figure of Figure 3 in the main paper.

\textbf{In Figure 3a}, for a given size of the dataset, i.e. for given number of objects, each query is a 1-in-K encoded vector of the object category, which means that the query dimensions grows with growing numbers of objects. This evaluates the representation in situations where objects are identified by a unique class index. Provided a sufficient number of gradient steps, we can see that the error stays low even with an increasing number of objects. We conjecture that the good performance of this setting is due to the growth in, both, query size and thus capacity of the model (as the query is the input to the model) as the number of objects grows.

\textbf{In Figure 3b}, the query vectors have a fixed dimension of $9$, equivalent to the dimension in the \textit{MultiON} benchmark. Queries are not 1-in-K encoded, but composed of randomly sampled values. Even though more gradient steps are helpful, the conclusion here is that increasing the number of objects has a negative impact on the mean error of the model. Unlike in (a), the number of parameters does not increase here with number of objects as queries have a fixed size. This is thus an illustration of the challenge to memorize an increasingly high number of objects with a fixed model capacity.

\textbf{Figure 3c} is a combination of (a) and (b) with query size increasing with number of objects and queries composed of random values (no one-hot vectors). The increase in model capacity with more objects seems again to be beneficial provided enough gradient steps. However, its is clear that the positions associated with random queries are more difficult to memorize than for one-hot queries. This emphasizes the importance of the chosen query representation when building query-able semantic implicit representations.

\section{Agent training details}
\noindent
All agents evaluated in this work are trained with Proximal Policy Optimization (PPO)~[44], following settings from previous work~[51, 34]. We provide a formulation of the reward function and the PPO hyper-parameters in the next sub-sections.
\myparagraph{Reward function} ---
\noindent
The reward function at time-step $t$ is composed of three terms,
\begin{equation}
R_{t}=\mathds{1}_{t}^{\text{reached}} \cdot R_{\text {goal }}+R_{\text {closer }}+R_{\text {time-penalty }}
\label{eq:eq_reward_ppo}
\end{equation}
where $\mathds{1}_{t}^{\text{reached}}$ is the indicator function that equals $1$ if the \textit{Found} action was called at time $t$ while being close enough to the target, and $0$ otherwise. $R_{\text {closer }}$ is a reward shaping term taking as value the decrease in geodesic distance to the next goal compared to previous timestep. Finally, $R_{\text {time-penalty }}$ is a negative slack reward to encourage taken paths to be as short as possible.

\myparagraph{PPO hyper-parameters}
\noindent
Table~\ref{table:ppo} presents the PPO hyper-parameter values used to train all agents in this work.

\begin{table}  \centering
  \caption{\textbf{PPO hyper-parameters:} Values for hyper-parameters used when training all agents in this work.}
  \label{table:ppo}
  \centering
  {\small
  \begin{tabular}{cc}
    \toprule
    Optimizer & Adam \\
    Adam eps & 1e-5 \\
    Learning rate & 2.5e-4 \\
    Linear learning rate decay & $\checkmark$ \\
    Number of epochs & 2 \\
    Number of parallel envs & 16 or 4 \\
    Number of mini batches & 4 or 1 \\
    Env. steps per update & 128 \\
    Clipping ratio & 0.2 \\
    Linear clip decay & $\checkmark$ \\
    Value loss coefficient & 0.5 \\
    Entropy coefficient & 0.01 \\
    Max Grad Norm & 0.2 \\
    GAE & $\checkmark$ \\
    GAE-$\lambda$ & 0.95 \\
    Discount factor & 0.99 \\
    Reward window size & 50 \\
    \bottomrule
  \end{tabular}
  }
\end{table}

\section{Amount of compute}
\noindent
Table~\ref{table:compute} shows the compute resources used to train, validate and test the different models involved in results presented in Table $1$ and $2$ in the main paper.

\begin{table*}  \centering
  \caption{\textbf{Amount of compute:} GPU days for runs involved in Tables $1$ and $2$ in the main paper.}
  \label{table:compute}
  \centering
  {\small
  \begin{tabular}{c ccc cc c ccccc}
    \toprule
    Type & $0-70$ & $30-70$ & $50-70$ & S & O & Nb episodes & Nb GPUs & GPU & GPU time & Nb runs & Tot. GPU time \\
    
    \multirow{3}{*}{Train} &&&&&&&& \\
     & $\checkmark$ & & & $-$ & $-$ && 1 & V100 & $5$d & $3$ & $15$d \\
     && $\checkmark$ & & $\checkmark$ & $-$ && 1 & V100 & $4.5$d & $3$ & $13.5$d \\
     &&& $\checkmark$ & $\checkmark$ & $\checkmark$ && 1 & V100 & $4$d & $3$ & $12$d \\
     
    \midrule
     
    Val &&&&&& $1000$ & $1$ & Titan X & $4$h & $90 (9*10)$ & $15$d \\
     
     \midrule
     
     Test &&&&&& $1000$ & $1$ & Titan X & $4$h & $9$ & $1.1$d \\
     
    \bottomrule
  \end{tabular}
  }
\end{table*}

\section{Limitations of the work}
\noindent
The proposed approach has the following limitations
\begin{itemize}
    \item Slower RL training compared to baseline agents. Even if reaching an average of $45$ fps with $2$ implicit representations updated with backpropagation at each agent step is already quite satisfying, we are still far from the $150$ fps when training $ProjNeuralMap$ or $200$ fps for $NoMap$.
    \item The \textit{Semantic Finder} can not deal with several instances of the same object type. This is not a problem when considering the \textit{MultiON} task, but will be addressed in future work.
    \item The \textit{Semantic Finder} only provides the position of an object of interest. Thus, it does not provide the agent with any information about how to reach the given target. Outputting a geodesic distance, and even a shortest path to the object would be interesting as future work.
    \item The current formulation of the uncertainty necessitates access to the full replay buffer of the episode. Future work will target estimating the uncertainty directly from the weights of the implicit representation.
\end{itemize}

\section{Leveraging environment
regularities and semantic priors in implicit representations}
\noindent
In this work, the weights of neural implicit representations are initialized from scratch at the beginning of each new episode and optimized in real time as the agent interacts with the environment. This is done in the same way as an explicit map would be updated on the fly during navigation. Training efficiency and the quality of the learnt representations are thus two important factors. Leveraging the knowledge about scene layout and semantic priors that was gained by each implicit representation to speed up the training of others and improving the quality of the provided mapping is thus a relevant future direction. Meta-learning better weight initializations, as was done in previous work (\textit{Sitzmann et al. MetaSDF: Meta-learning Signed Distance Functions, NeurIPS 2020)}, or having a common backbone followed by randomly initialized layers for new episodes are two promising directions.

\section{Importance of Fourier features}
\begin{figure*}
    \centering
    \includegraphics[width=0.7\textwidth]{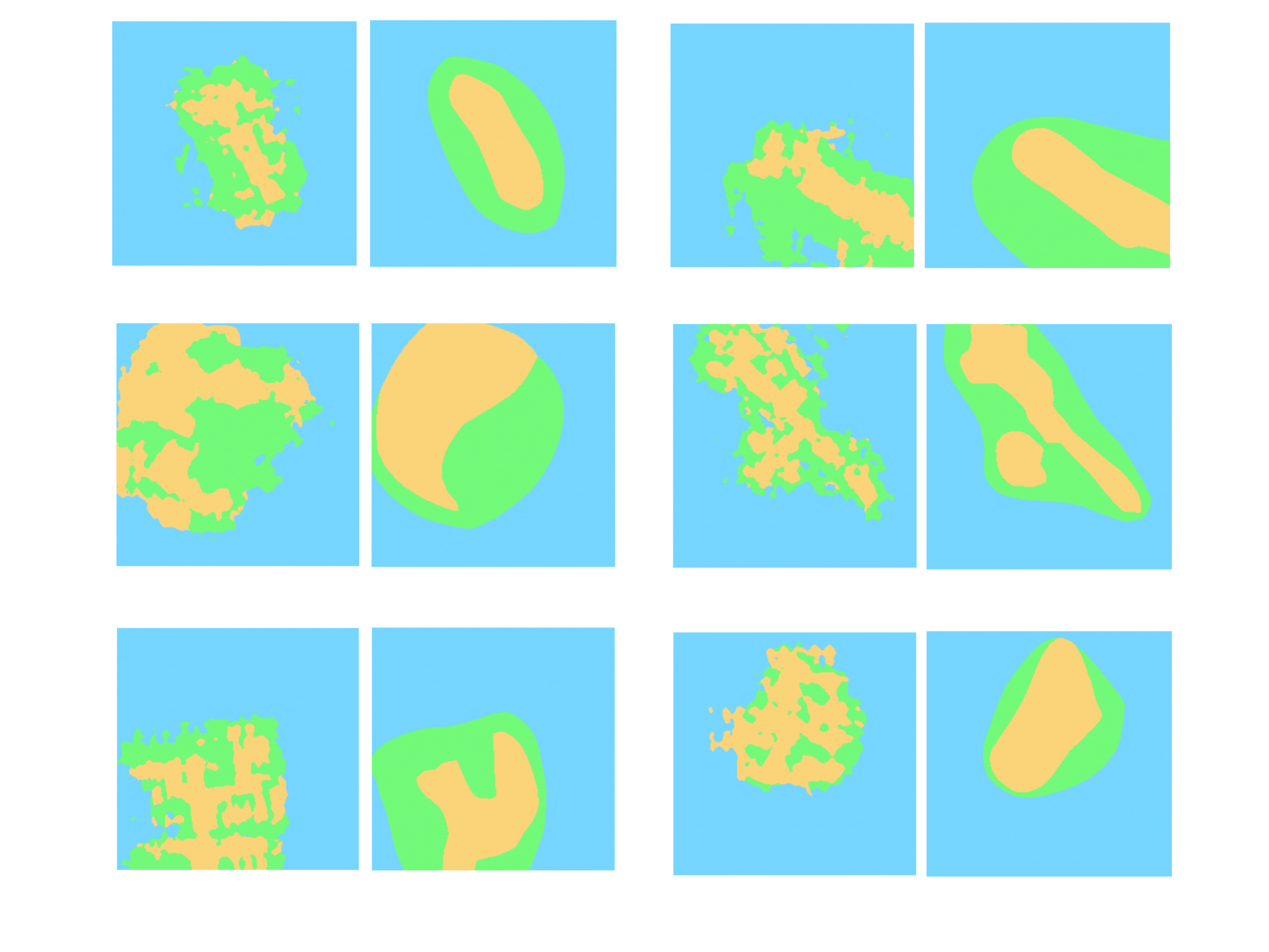}
    \caption{Comparison of top-down maps obtained by querying the \textit{Occupancy and Exploration Implicit Representation} trained with (left) and without (right) Fourier features.}
    \label{fig:fourier_features}
\end{figure*}

\noindent
Figure~\ref{fig:fourier_features} compares the top-down map obtained after querying the \textit{Occupancy and Exploration Implicit Representation} trained with and without Fourier features. On each plot, the left and right maps respectively show the impact of using and not using Fourier features. Without the latter, no detail about the environment layout can be reconstructed. This corroborates findings also reported in other literature on implicit representations and coordinate networks, e.g. \textit{(Mildenhall et al., NeRF: Representing Scenes as Neural Radiance Fields for View Synthesis, ECCV 2020)}

\section{Importance of Occupancy and Exploration information}
\noindent
In an instance based post-hoc analysis, we attempt to visualize the way how the implicit \textit{Occupancy and Exploration Representation} is used by the agent with two different examples.
\subsection{Occupancy information}
\begin{figure}
    \centering
    \includegraphics[width=0.53\textwidth]{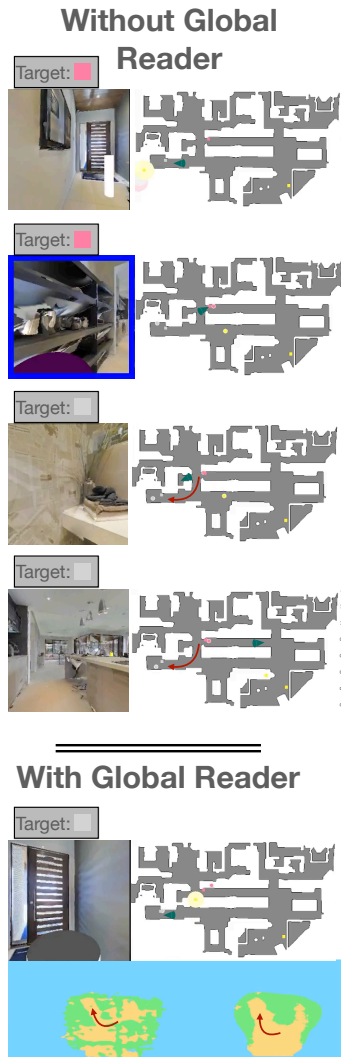}
    \caption{\textbf{Importance of occupancy information}: Episode example from the \textit{MultiON} 2021 challenge minival set where an agent without the proposed \textit{Occupancy and Exploration Implicit Representation} fails to find the white target. Another agent equipped with the representation containing occupancy information finds the target. Some information about the path to reach it (indicated with the red arrow) is indeed contained in the map. The green cone corresponds to the agent's position.}
    \label{fig:importance_occupancy}
\end{figure}

\noindent
Figure~\ref{fig:importance_occupancy} shows an example episode from the \textit{MultiON} 2021 challenge minival set targeted by two different agent variants.

\myparagraph{Without the Occupancy and Exploration Representation} the agent fails to find a target that was already observed in the past: at the beginning of the episode, the agent observes the white cylinder, while the current target is the pink object. The agent thus explores the environment, finds it and properly calls the \textit{Found} action. The next target is then the white object. As can be seen, the \textit{Semantic Finder} properly locates the target, since it had been previously observed. However, the agent fails to backtrack, first entering a room without finding a path to the goal and then going in the wrong direction. It finally calls the \textit{Found} action while not being close to the white target. 

\myparagraph{The full agent} also having access to the  \textit{Occupancy and Exploration Implicit Representation} succeeds in reaching the white target (after finding the pink target, not shown on the Figure). Moreover, in visualizations of, both, the implicit representation $f_o$ and the reconstruction from the latent representation extracted by the Global Reader $r$, we can see the path to the goal (visualized as a red arrow) marked as explored area. All the information for the agent to correctly navigate towards the white object is thus contained in the implicit map and its global summary vector.

This analysis cannot corroborate that the agent indeed used the representation as explained; however, we can at least provide evidence that the (successful) full agent had access to information, which was crucial to solve a task, on which the baseline agent failed.

\subsection{Exploration information}
\begin{figure}
    \centering
    \includegraphics[width=0.45\textwidth]{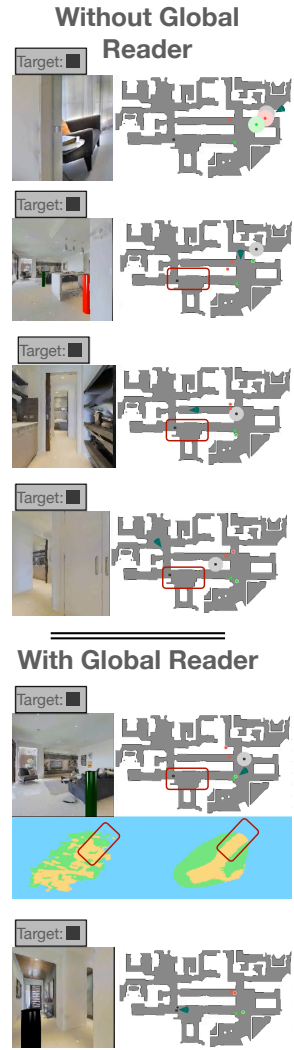}
    \caption{\textbf{Importance of exploration information}: Episode example from the \textit{MultiON} 2021 challenge minival set where an agent without the proposed \textit{Occupancy and Exploration Implicit Representation} fails to find the black target. Another agent equipped with the representation containing occupancy information finds the target. Some information about the area to explore (indicated with the red shape) is indeed contained in the map. The green cone corresponds to the agent's position.}
    \label{fig:importance_exploration}
\end{figure}

\noindent
Figure~\ref{fig:importance_exploration} shows another example episode taken from the \textit{MultiON} 2021 challenge minival set.

\myparagraph{Without the Occupancy and Exploration Representation} the agent fails to explore the scene to find a target: the first target object is black. To this end, the agent must explore the environment. It starts to explore but misses a part of the scene (containing the black target), which is never explored for the rest of the episode. It finally calls the \textit{Found} action far away from the black target.

\myparagraph{The full agent} also having access to the  \textit{Occupancy and Exploration Implicit Representation} succeeds in finding the black target. It successfully explores the part of the scene that contains the object. In visualizations of, both, the implicit representation $f_o$ and the reconstruction from the latent vector extracted by the Global Reader $r$, we can see that the area to observe (visualized by a red shape) is at the frontier between  explored and unexplored parts of the scene. This information can thus guide the agent to move towards the unexplored area.

\end{document}